\PassOptionsToPackage{table,xcdraw,dvipsnames}{xcolor}

\documentclass{article}

\PassOptionsToPackage{numbers, compress}{natbib}

    \usepackage[final]{neurips_2024}

\usepackage[utf8]{inputenc} 
\usepackage[T1]{fontenc}    
\usepackage{hyperref}       
\hypersetup{
    colorlinks=true,
    linkcolor=BlueViolet,
    citecolor=teal
}
\usepackage{url}            
\usepackage{booktabs}       
\usepackage{amsfonts}       
\usepackage{nicefrac}       
\usepackage{microtype}      
\usepackage{xcolor}         

\usepackage{makecell}
\usepackage{amsthm}
\usepackage{bm}
\usepackage[table,xcdraw]{xcolor}
\usepackage{multicol}
\usepackage{colortbl}
\usepackage{amsmath}
\usepackage{cleveref}
\usepackage{multirow}
\usepackage{graphicx}
\usepackage{subcaption}
\usepackage{enumitem}
\usepackage{bbding}
\usepackage{color}

\usepackage{algorithm}
\usepackage{listings}
\usepackage{natbib}

\usepackage{mdframed}
\usepackage{amsmath,bm,bbm}
\usepackage{footnote}
\definecolor{baselinecolor}{gray}{.9}

\usepackage{etoolbox}
\makeatletter
\AfterEndEnvironment{algorithm}{\let\@algcomment\relax}
\AtEndEnvironment{algorithm}{\kern2pt\hrule\relax\vskip3pt\@algcomment}
\let\@algcomment\relax
\newcommand\algcomment[1]{\def\@algcomment{\footnotesize#1}}
\renewcommand\fs@ruled{\def\@fs@cfont{\bfseries}\let\@fs@capt\floatc@ruled
  \def\@fs@pre{\hrule height.8pt depth0pt \kern2pt}%
  \def\@fs@post{}%
  \def\@fs@mid{\kern2pt\hrule\kern2pt}%
  \let\@fs@iftopcapt\iftrue}
\makeatother
\definecolor{tabhighlight}{HTML}{e5e5e5}
\definecolor{tabhighlight2}{HTML}{e8e8e8}
\newcommand{\tabstyle}[1]{
  \setlength{\tabcolsep}{#1}
  \renewcommand{\arraystretch}{\tableCellHeight}
  \centering
  \small
}
\newcommand{\para}[1]{
  \noindent\textbf{#1}
}

\definecolor{ForestGreen}{RGB}{34,139,34}

\newcommand{\ie}{\emph{i.e.},~}

\renewcommand{\paragraph}[1]{\medskip\noindent\textbf{#1.~}}

\theoremstyle{plain}
\newmdtheoremenv{corollary}{Corollary}

\newtheorem{prop}{Proposition}
\newtheorem{assump}{Assumption}
\newtheorem{definition}{Definition}

\newmdtheoremenv[linewidth=0pt,innerleftmargin=4pt,innerrightmargin=4pt]{lemma}{Lemma}%

\usepackage{tikz}
\newcommand{\tableCellHeight}{1}
\newcommand*{\circled}[1]{\lower.7ex\hbox{\tikz\draw (0pt, 0pt)%
    circle (.5em) node {\makebox[1em][c]{\small #1}};}}

\newcommand{\myref}[1]{Eq.\eqref{#1}}
\newcommand{\modelname}{{BoostAdapter}}

\title{BoostAdapter: Improving Vision-Language Test-Time Adaptation via Regional Bootstrapping}

\author{%
   Taolin Zhang$^{1}$ \quad  
   Jinpeng Wang $^{1}$ \quad 
   Hang Guo $^{1}$ \\
    \textbf{ Tao Dai\thanks{Correspongding author: Tao Dai (daitao.edu@gmail.com)} $^{\ 2}$ \quad  
    Bin Chen   $^{3}$ \quad 
    Shu-tao Xia $^{1,4}$ } \vspace{0.1cm} \\
 $^1$ Tsinghua University \quad
$^2$ Shenzhen University \\
$^3$ Harbin Institute of Technology \quad 
$^4$ PengCheng Laboratory \\ 
  \url{https://github.com/taolinzhang/BoostAdapter}
}

\begin{document}

\maketitle
\begin{abstract}
Adaptation of 
pretrained vision-language models such as CLIP to various downstream tasks have raised great interest in recent researches. 
Previous works have proposed a variety of test-time adaptation (TTA) methods to achieve strong generalization without any knowledge of the target domain. 
However, existing training-required TTA approaches like TPT necessitate entropy minimization that involves large computational overhead, while training-free methods like TDA overlook the potential for information mining from the test samples themselves.
In this paper, we break down the design of existing popular training-required and training-free TTA methods and bridge the gap between them within our framework.
Specifically, we maintain a light-weight key-value memory for feature retrieval from  instance-agnostic historical samples and instance-aware boosting samples. 
The historical samples are filtered from the testing data stream and serve to extract useful information from the target distribution, while the boosting samples are drawn from regional bootstrapping and capture the knowledge of the test sample itself.
We theoretically justify the rationality behind our method and empirically verify its effectiveness on both the out-of-distribution and the cross-domain datasets, showcasing its applicability in  real-world situations.  
\end{abstract}
\section{Introduction}
\label{sec: introduction}

Vision Language models \cite{yang2022vision, jia2021scaling, li2021align,li2022blip,li2023blip, gao2024energy} have shown incredible performance in downstream vision tasks \cite{bommasani2021opportunities}, such as classification \cite{lu2022prompt,zhou2022coop,zhou2022cocoop, gao2024clip}, generation \cite{kumari2023multi, rombach2022high, guo2024refir} and recognition \cite{wang2023clip, wasim2023vita}. 
Among these models, CLIP \cite{radford2021learning} has been trained with large-scale noisy image-text pairs and can generalize well in zero-shot recognition tasks. The key idea behind CLIP is modality alignment during training and similarity comparison during testing for classification. However, CLIP suffers from domain shift problems during test-time inference. In the presence of out-of-distribution issues \cite{liang2017enhancing, wang2020tent, hendrycks2019benchmarking} that commonly appear in real-world scenarios, CLIP may fail to effectively align the feature across modalities, leading to performance degradation.

\begin{figure}[!t]
    \centering    \includegraphics[width=\textwidth]{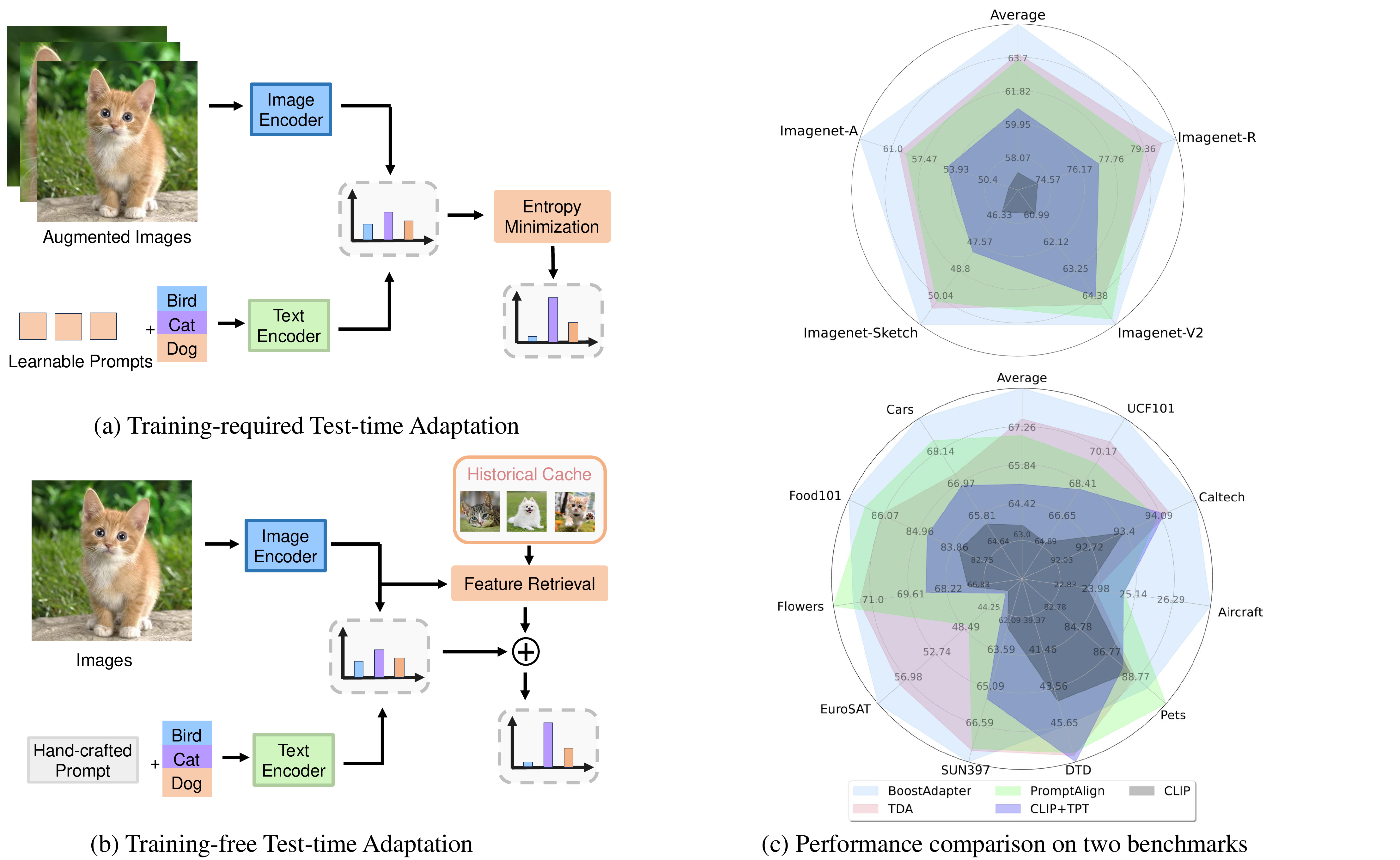}
    \caption{
        \small
        (a) Existing training-required TTA methods utilize self-supervised objective like entropy minimization for better generalization.
        (b) Existing training-free TTA methods perform feature retrieval on the historical samples to adjust the model prediction.
        (c) Performance comparison on the Out-of-Distribution benchmark and Cross-Datasets benchmark. 
    }
    \label{fig:intro}
    \vspace{-1em}
\end{figure}

Test-time adaptation (TTA) has been widely explored in recent approaches \cite{wang2020tent, iwasawa2021test, shu2022test, karmanov2024efficient} to mitigate misalignment issues and improve performance in downstream tasks. Current mainstream TTA methods can be divided into training-required methods and training-free methods, as depicted in Figure. \ref{fig:intro}a and Figure. \ref{fig:intro}b.
Training-required approaches \cite{wang2020tent, shu2022test,samadh2023align} adjust model parameters or learnable prompts based on self-supervised objectives like entropy and increase the prediction confidence of model for distribution adaptation. TPT \cite{shu2022test} applies entropy minimization to the vision-language model first.  Furthermore, inspired by consistency regularization, TPT performs information mining from the test sample itself by random regional cropping in a self-bootstrapping style. However, training-required methods require gradient descent that is time-consuming with large training overhead, which prevents them from being applied in computationally limited situations.
Training-free approaches \cite{iwasawa2021test, zhang2022tip, karmanov2024efficient} utilize memory networks, cache, or prototypes to store information regarding target samples and distributions, which is then used to adaptively modify the model's prediction. For example, TDA \cite{karmanov2024efficient} leverages historical samples from the test data steam to build a dynamic key-value cache. It updates the prior knowledge encoded in CLIP through feature retrieval and output prediction based on the similarity between the test sample and the high-quality data stored in the memory bank.
However, existing training-free approaches only consider interaction with other historical samples in the cache and do not effectively exploit the information within the test sample itself. This limitation prevents them from performing well especially in tasks that require fine-grained information. 

Both of these approaches demonstrate excellent performance in enhancing the robustness of vision-language models to unknown distributions. However, the connection between them remains unclear. In this paper, we aim to answer three questions: (1) How are training-required methods like TPT and training-free methods like TDA connected? (2) How can we combine these two methods based on their shared nature? (3) Does vision-language models benefit from the combination of these methods?

In order to answer these questions, we first consider that the augmented images of test samples form a regional bootstrapping distribution of the original data. By filtering out the noisy augmentations based on mutual information with the predefined CLIP text embedding clusters, we can obtain a \textbf{boosting distribution} from which high-quality samples close to the target clusters can be drawn. 
Based on this, we delve into the connection between the target operations over the boosting distribution, \ie cross-entropy optimizations and cache classifier, which reveals the shared nature between entropy-based and cache-based methods. Specifically, we pinpoint that with the samples derived from the bootstrapping distribution, entropy minimization over them performs equivalently to feature retrieval from the cache consisting of them.
Motivated by this analysis, we propose a brand-new adaptation strategy, dubbed \textbf{BoostAdapter}, to improve training-free adapters by incorporating the samples derived from the boosting distribution to the memory bank. 
Particularly, the cache in BoostAdapter consists of instance-agnostic historical samples filtered from the test data stream, along with instance-aware boosting samples generated through regional bootstrapping from the sample itself. The interactions between intra-sample and cross-sample operations make BoostAdapter effective and efficient by incorporating the idea of information mining from training-required methods while maintaining the efficiency of training-free methods. Theoretical analyses and empirical results are also provided to validate the effectiveness of BoostAdapter.

To summarize, we make the following contributions in this paper.
\setlist{nolistsep}
\begin{itemize}[leftmargin=1.5em]
    \item We first discuss the relationship between training-required and training-free methods in test-time adaptation and establish connections between them.   
    \item We propose BoostAdapter, a brand new adaptation strategy in test-time adaptation of vision-language models, which improves training-free adapters by introducing high-quality samples from regional bootstrapping into the memory.
    \item We theoretically derive target domain error bound of BoostAdapter and shows that BoostAdapter benefit from incorporating self-bootstrapping data. 
    \item Extensive experiments conducted over two benchmark demonstrate the superior performance of BoostAdapter under test-time adaptation settings.
\end{itemize}

\section{Related Works}
\label{sec:related_work}
\para{Vision-Language Models} have shown remarkable potential in generalization by contrastive pre-training over amounts of text-image pairs \cite{jia2021scaling, radford2021learning,li2022blip, li2023blip} . One typical work is CLIP \cite{radford2021learning}, which benefits from the alignment of 400 million curated image-text pairs and predicts the most relevant text description for a given image based on cosine similarity. 
Adapting CLIP to the downstream applications has attracted much attention and has been widely explored in recent approaches \cite{zhou2022coop, zhou2022cocoop,zhang2022tip, li2024graphadapter, zhu2023not, lu2023beyond}. CoOp \cite{zhou2022coop} introduces learnable prompts \cite{lester2021power, zha2023instance, yang2024not, liu2021p} and CoCoOp \cite{zhou2022cocoop} conditions the text prompts on image embedding for better generalization. Maple \cite{khattakMaPLe} performs prompting for both vision and language branches and improves the alignment of the embedding between modalities. 
These approaches have demonstrated significant performance enhancements, but they still require  few training data from the target domain. In contrast, we focus on test-time adaptation where there is no information about the target distribution and aim to generalize the model to any unknown scenarios.

\para{Training-required Test-time Adaptation} updates partial wights of the model like prompts \cite{shu2022test,samadh2023align} or BN layer \cite{wang2020tent} with self-supervised objectives that benefit the downstream tasks without requiring additional training data. Tent \cite{wang2020tent} reduces generalization error on shifted data by test-time entropy minimization. 
For vision-language models, Test-time prompt tuning (TPT) \cite{shu2022test} is a method that dynamically optimizes prompts during the testing phase, enhancing the model's zero-shot generalization ability. Specifically, TPT generates multiple augmented views of the test sample and then minimizes the entropy of the model's output logits across them to ensure consistent prediction. 
Recently, many works built upon TPT have been proposed to further enhance the performance of vision-language models.  
Particularly, DiffTPT \cite{feng2023diverse} leverages the power of diffusion models to generate semantically consistent augmented images for entropy minimization. 
PromptAlign \cite{samadh2023align}  bridges the gap between the test sample and source distribution by aligning token statistics, including mean and variance.
Nevertheless, these approaches require gradient descent over the augmented images, which is computationally expensive and time-consuming.

\para{Training-free Test-time Adaptation} applies cache model or prototypes to make prediction of test samples in a non-parametric manner \cite{iwasawa2021test,karmanov2024efficient, zhang2023adanpc}. T3A \cite{iwasawa2021test} utilizes prototypes as downstream classifiers and dynamically adjusts the weights. AdaNPC \cite{zhang2023adanpc} leverages the data from the source domain to address the issues of computation overhead and domain forgetting. 
For vision-language models, TDA \cite{karmanov2024efficient} introduces both positive cache and negative cache to obtain high-quality test samples from the target domain. However, these methods only consider inter-sample interactions and may fail to generalize well when the downstream tasks require fine-grained knowledge or there is insufficient similarity across samples. 

\section{Methodology}
\label{sec:method}
\subsection{Preliminary}
\para{Problem setting.} 
We begin by introducing the basic notations in test-time adaptation. We consider binary classification for simplicity and the theory can be easily extended to multi-classifications settings.
Let $p_t{(x,y)}$ denotes the joint distribution of image and labels in the target distribution, and we simply assume that samples $\{(x_i,y_i)\}_{i=1}^n$ are drawn i.i.d. from the distribution with $y_i$ represents the one-hot label.
\begin{definition}
(\textbf{Classification error.}) Given $f$ as a binary classification function. The error incurred by hypothesis $f\in\mathcal{H}:\mathcal{X}\rightarrow\{0,1\}$ under the distribution $p_t{(x,y)}$ can be defined as 
\begin{small}
\begin{equation}
\begin{aligned}
\epsilon({f})&=\mathbb{E}_{p_t{(x,y)}}[{f}(x)\neq y] = \mathbb{E}_{p_t{(x,y)}}[|{f}(x)-y|], 
\end{aligned}
    \label{equ:excess}
\end{equation}
\end{small}
the last equality holds in a binary classification setting. 
\end{definition}

\begin{definition}
    (\textbf{Excess error}.) Given the Bayes classifier under distribution $p_t{(x)}$: $f^*(x)= \mathbb{I}\{f(x)\geq 1/2\}$ and the optimal classfier $f^*$, the excess error of $f$ is defined as
\begin{equation}
\begin{aligned}
\mathcal{E}(f)&=\epsilon(f)-\epsilon(f^*)
=2\mathbb{E}_{x\sim p_t(x)}\left[\left|f(x)-\frac{1}{2}\right|\mathbb{I}\{f(x)\neq f^*(x)\}\right] 
\end{aligned}
\end{equation}
\end{definition}

\para{CLIP classifier}
Let $g$ be the image encoder of CLIP, $C$ be the feature dimension, $N$ denotes the number of categories, $w_i \in R^{C}$ represents the $i_{th}$ text embedding cluster.
Considering normalized embedding $w$ and $g(x)$, we can derive a simplified version of the output of CLIP for class $i$:
\begin{equation}
    Z_i={w}_{{i}}^Tg(x).
    \label{equ:clip}
\end{equation}
And we denote the output logits as $\boldsymbol{p}(x)=[Z_1, Z_2, ..., Z_N]\in R^{N}$.  

\para{Cache classifier}
Given an unseen sample $x$, encoder $g$ with dimensional $C$, cache size $K$ and number of categories $N$, the cache classfier conduct feature retrieval based on the similarity with the data $\{(x_i, y_i)\}_{i=1}^k$ in the cache. The predictions based on Tip-Adapter \cite{zhang2022tip} are as follows:
\begin{equation}
    \boldsymbol{p_{cache}}({x})=A\left(g(x) G_{cache}^T\right)Y,
    \label{equ:cache}
\end{equation}
where $A(z) = \alpha \text{exp}(-\beta(1 - z))$ denotes a scaling function with a weighting factor $\alpha$ and a smoothing scalar $\beta$, $G_{cache}\in R^{K\times C}$ represents the feature of $K$ samples $\{x_i\}_{i=1}^K$ in the cache and $Y\in R^{K\times N}$ is the corresponding labels $\{y_i\}_{i=1}^K$.
Considering the number of samples in class $y_i$, We can also derive a simplified version of \myref{equ:cache}   as follows, by ignoring the scaling function and adopting an instance-wise computation style:
\begin{equation}
    \boldsymbol{p_{cache}}({x})=\sum_{i=1}^{k}\alpha_i \left[g(x_i)^T g(x)\right]y_i, 
    \label{equ:cache_simple}
\end{equation}
where $\alpha_i = \frac{1}{n_{y_i}}$ for class balance or $\alpha_i = \frac{1}{\sum_{j=1}^k[g(x_j)^T g(x)]}$ for normalization across all the samples.

\begin{figure}[!t]
    \centering    \includegraphics[width=0.8\textwidth]{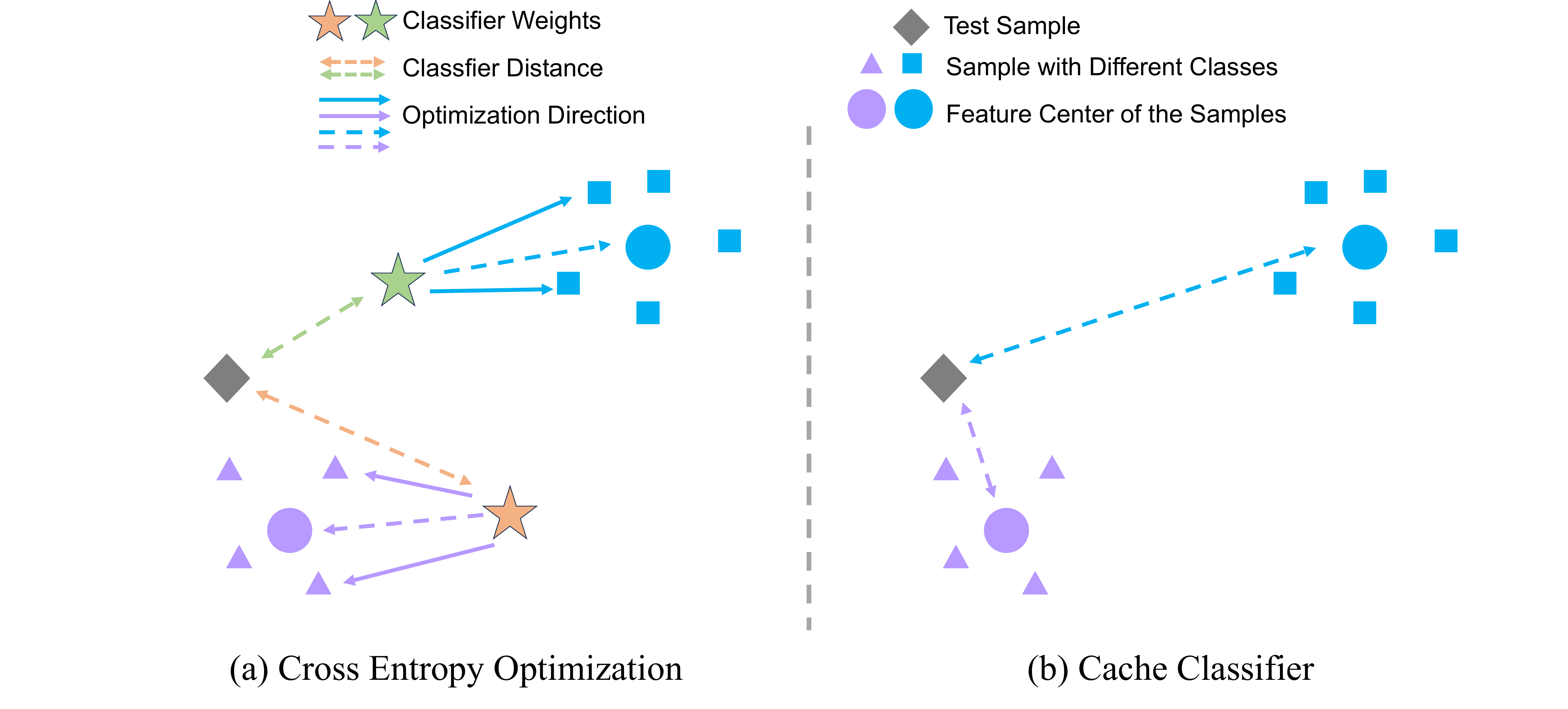}
    \vspace{-5pt}
    \caption{
        \small
        Connection between cross-entropy optimization and cache classifier over well-clustered samples with a frozen feature encoder. With optimization of cross-entropy, samples will pull the classifier weights closer of the same class while pushing them away from different class weights. Since the feature space is well-clustered, the classifier weights will ultimately converge near the feature center of the samples. Finally, the optimal classifier achieved through cross-entropy minimization will exhibit similar behavior with the cache classifier.  
    }
    \label{fig:eq}
\end{figure}
\subsection{A Closer Look at Entropy-based and Cache-based Methods}
We start with analyzing the filtering operation of augmentated images in TPT. Pseudo-labels tends to be noisy in the test time, and entropy can serve as a confidence metric to identify trustworthy samples among augmented views \cite{wang2020tent, shu2022test, niu2023towards}. These high-quality samples can be considered drawn i.i.d. from the so-called boosting distribution as defined below. 
\begin{definition}
(\textbf{Boosting Distribution}.) Given a test sample from target distribution $x\sim p_t(x)$, let $H(\cdot)$ be the entropy measuring function and $Aug(\cdot)$ be the regional augmentation. By filtering noisy samples based on thresthould $\tau$, we have the following property of boosting distribution $p_b(x)$:
\begin{equation}
 \hat{x} \sim p_b(x) \to \{\hat{x}=Aug(x)\wedge H(\boldsymbol{p}(x)) \le \tau \} 
\end{equation}  
\end{definition}
We also terms the samples from the boosting distribution as \textbf{boosting samples}.
Then we can connect entropy-based methods and cache classfier by the following proposition:
\begin{prop} 
    (Informal)
    Given $n$ samples $\{(x_i, y_i)\}_{i=1}^{n}$ with a freeze encoder $g$ that effectively performing feature clustering with respect to labels, the gradient descent optimization direction of the classifier's weights based on cross-entropy generally tends towards making predictions using the cache classifier with class balance weights defined in \ref{equ:cache_simple} on these samples.
    \label{prop:eq}
\end{prop}

An intuitive illustration of Proposition \ref{prop:eq} is depicted in Figure \ref{fig:eq}, where the weights of optimal classfier behave like the feature centers across different classes with of the well-clusterd samples.
Revisiting the entropy-based method TPT, when provided with high-quality boosting samples with low entropy drawn from the boosting distribution, the objective function of entropy minimization optimizes in a manner similar to conducting cross-entropy optimization over the pseudo-labels.
According to Proposition \ref{prop:eq}, TPT performs similarly to the cache-based methods with a cache comprising the same boosting samples from the boosting distribution.

\begin{figure}[!t]
    \centering    \includegraphics[width=0.9\textwidth]{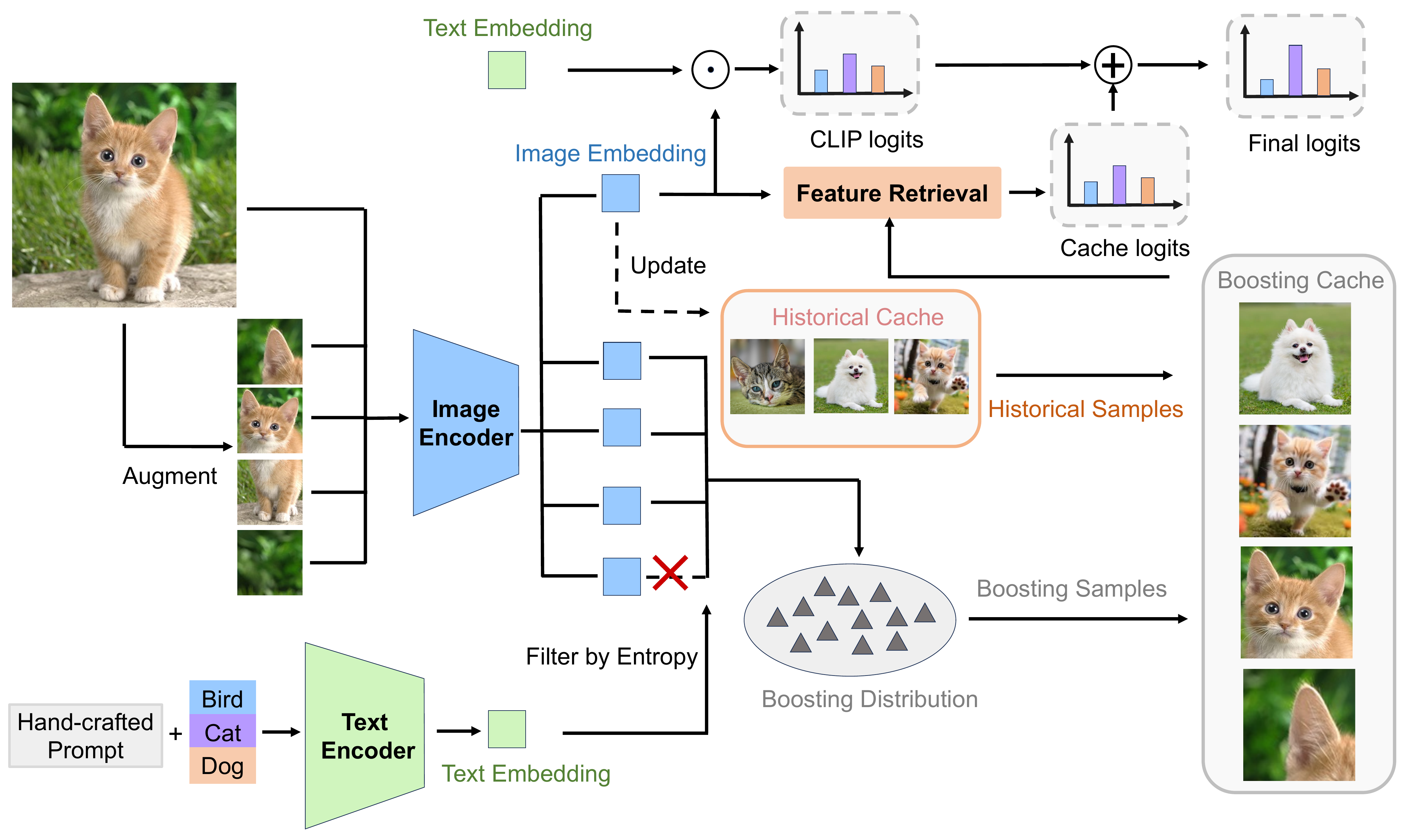}
    \vspace{-5pt}
    \caption{\small  
    \textbf{Overall architecture of BoostAdapter.} BoostAdapter leverages knowledge from the target domain and employs self-bootstrapping with historical and boosting samples in the boosting cache, respectively.
    }
\end{figure}

\subsection{Boosting your Training-free Adapters}
Existing cache-based methods store historical test samples only as useful information for prediction. In light of the analysis above, we can integrate the idea behind TPT into these training-free adapters by incorporating boosting samples into the memory bank. 
In particular, each sample can participate in both inter-sample and intra-sample interactions with the instance-agnostic historical samples and the instance-aware boosting samples in the cache, respectively.

Specifically, with $k_t$ selected historical samples and $k_b$ selected boosting samples to comprise the cache, we extend the classifier defined in \myref{equ:cache} and formulate our BoostAdapter as follows:
\begin{equation}
    \boldsymbol{p_{boost}}({x})=A\left(g(x) \tilde{G}_{cache}^T\right)\tilde{Y},
    \label{equ:boostadapter}
\end{equation}
where $A$ is the same scaling function defined in \myref{equ:cache}, $\tilde{G}_{cache}\in R^{(k_t+k_b)\times C}$ denotes the features of the combination of both the historical and boosting samples, and $\tilde{Y}\in R^{(k_t+k_b)\times N}$ is the label. 

Since we do not have access to the labels of the test samples, we generate one-hot pseudo-labels for them using argmax operations. However, these pseudo-labels tend to be noisy in the target domain. Therefore, we apply filtering based on entropy thresholds on the test data stream following \cite{shu2022test} to obtain trustworthy historical samples. We employ a similar operation to select boosting samples from multiple augmented views of the current sample.
In practice, we dynamically adapt the entropy thresholds $\tau$ for each test sample, with a fixed percentile $p$. The cache continuously updates with lower entropy historical samples from the test data stream, while the current test sample augments the cache with self-boosting samples and forms an independent cache that only affects its own prediction.
Additionally, to maintain diversity while considering the relevance to each test sample, we set a maximum shot capacity for each class $k$ in the cache. This means that samples in the cache will be replaced by a lower-entropy historical sample or boosting sample when necessary. 

An important issue is whether introducing boosting samples brings improvements to the training-free adapters. We will first make some necessary assumptions and then theoretically verify the effectiveness in reducing target error by incorporating samples from the boosting distribution.
\begin{assump}
    \textit{(Strong Density Condition)} For any test sample $x_0$ in the target distribution $x_0 \sim p_t(x)$ and the boosting distribution $p_b(x_0)$, given positive lower bound $m$ and upper bound $M$, positive scaling constant $c_t$ and $c_{b}$, the radius bound $R>0$, and $\mathcal{B}(x,r)=\{x':\parallel x'-x\parallel\leq r\}$ is the ball centered on $x$ with radius $r$. We assume $p_t(x)$ and $p_b(x_0)$ are absolutely continuous with respect to the Lebesgue measure in $\mathbb{R}^d$. For $r\in(0,R]$, we assume
    \begin{equation}
    \left\{
    \begin{aligned}
     & \lambda[p_t(x)\cap \mathcal{B}(x_0,r)] \geq c_t\lambda[\mathcal{B}(x_0,r)]\\
     & \lambda[p_b(x_0)\cap \mathcal{B}(x_0,r)] \geq c_b \lambda[\mathcal{B}(x_0,r)]\\
     &  m<\frac{d p_t(x)}{d \lambda} < M; m<\frac{d p_b(x)}{d \lambda} < M,
    \end{aligned}
    \right.
    \end{equation}
    where $\lambda$ is the Lebesgue measure in Euclidean space.
    \label{assump1}
\end{assump}

\begin{assump}
    (L-Lipschitz Condition) Let $f$ be the classification function and $L$ be a positive constant. For all feasible $x,x^{\prime}$ we have $|f(x)-f(x')|\leq L\parallel x-x' \parallel$.
    \label{assump2}
\end{assump}

\begin{assump}
    (Low Noise Condition). Let $\beta,C_\beta$ be positive constants and we assume $p_t(x)$ satisfies $P_{x\sim p_t(x)}\left(\left| f(x)-\frac{1}{2} \right|<t\right)\leq C_\beta t^\beta$ for all $t>0$.
    \label{define_noise}
\end{assump}

\para{Remark} Assumption \ref{assump1} intuitively ensures that for any test sample, there is a surrounding neighborhood with a significant presence of samples from the target domain and the boosting distribution. 
More importantly, for a specific sample $x_0$, boosting samples $x\sim p_b(x_0)$ should be closer to $x_0$ than other samples $x\sim p_t(x)$ from the target domain, \ie generally, we have $c_t \le c_b$. 
Assumption \ref{assump2} and \ref{define_noise} describe the smoothness of functions and imply a high level of confidence in predictions around the threshold, respectively.

\begin{prop}
\textbf{(Historical Cache reduce Emperical Risk)}
Given $f$ as the training-free classfier consisting of historical samples only defined by \myref{equ:cache}. Let $n_t$ to be the number of confident previously predicted samples in the target domain and $k_t$ as the number of historical samples in the cache, with assumptions 1-3, the following results hold with high-probability for large enough $k_t$ and $n_t$. 
\begin{equation}
\begin{aligned}
\mathcal{E}(f)&\leq \mathcal{O}\left(\left(\frac{1}{k_t}\right)^{1/4}+  \left(\frac{k_t}{c_t n_t}\right)^{1 /{d}} \right)^{{1+\beta}}
\end{aligned}
\end{equation}

\label{prop:knn_his}
\end{prop}

\begin{prop}
\textbf{(Historical Cache benefits from Boosting Samples)}
Let $n_t$ to be all confident previously predicted samples in the target domain and $n_b$ be the number of boosting samples that are drawn from the boosting distribution. 
Given $k_t$ and $k_b$ to be the number of historical samples and the number of boosting samples to be selected as the nearest neighbors stored in the cache, respectively. Let $w_{ti}$ and $w_{bi}$ be the weights defined in \myref{equ:cache_simple} of the historical samples and boosting samples.
We have the following bound for the empirical risk of the cache classfier defined in \ref{equ:boostadapter}. 

\begin{equation}
\begin{aligned}
\mathcal{E}(f)\leq \mathcal{O}\left(\left(\frac{1}{k_t+k_b}\right)^{1/4}+\sum_{i=1}^{k_t} {w_{ti}} \left(\frac{k_t}{c_t n_t}\right)^{1 /{d}} +\right. 
\left.\sum_{i=1}^{k_b} {w_{bi}} \left(\frac{k_b}{c_b n_b}\right)^{1 /{d}}   \right)^{{1+\beta}} .
\end{aligned}
\end{equation}
\label{prop:knn_his_boost}
\end{prop}

\para{Remark} Proposition \ref{prop:knn_his} provides a guarantee of the effectiveness of selecting  $k_t$ out of $n_t$ historical samples to comprise the cache. The empirical risk is quite small when $n_t\to \infty$ since the cache captures the full information of the target domain.
Proposition \ref{prop:knn_his_boost} demonstrates that the historical cache can further reduce empirical risk by incorporating $k_b$ boosting samples.

\label{subsec:overview}
\section{Experiments}
\label{sec:experiments}

\subsection{Experimental Setup}
\label{subsec:setup}
\para{Datasets}
Following the setting in TPT \cite{shu2022test}, we conduct experiments on both Out-of-Distribution (OOD) benchmark and Cross-Domain benchmark. 
The OOD benchmark evaluates the model’s robustness to natural distribution shifts on 4 ImageNet \cite{deng2009imagenet}
Variants, including ImageNetV2 \cite{recht2019imagenet}, ImageNet-Sketch \cite{wang2019learning}, ImageNet-A \cite{hendrycks2021natural} and ImageNet-R \cite{hendrycks2021many}.  
We evaluate the transferring performance on 11 datasets in the Cross-Domain benchmark: Aircraft \cite{maji2013fine},
Caltech101 \cite{fei2004learning}, Cars \cite{krause20133d}, DTD \cite{cimpoi2014describing}, EuroSAT \cite{helber2019eurosat}, Flower102 \cite{nilsback2008automated}, Food101 \cite{bossard2014food}, Pets \cite{parkhi2012cats}, SUN397 \cite{xiao2010sun},and UCF101 \cite{soomro2012dataset}. We follow the split in \cite{zhou2022coop} and report the top-1 accuracy. The error bound are also provided.

\para{Implementation details}
\label{sec:details}
We utilize a pre-trained ViT-B/16 of CLIP as the foundation model. In test-time adaptation, the batch size is set to be 1. We search for the optimal shot capacity to balance diversity and relevance of samples. For boosting samples, we utilize random crop and then random horizontal flip as augmentations. Moreover, we empirically set the entropy threshold percentile to  $p=0.1$ and filter 64 augmented views based on random cropping to obtain the boosting samples. and filter 64 augmented views to obtain the boosting samples. The top-1 accuracy and the error bound is reported on the test sets. All our experiments are conducted with a Nvidia 3090 24GB GPU.

\begin{table}[!t]
    \caption{\textnormal{\textbf{Full results on the OOD benchmark with ViT-B/16 backbone.} } We report  top-1 accuracy and ``Average" is calculated by taking the mean accuracy across all four OOD datasets.} 
    \small \centering
 \setlength{\tabcolsep}{8pt}
    \resizebox{0.9\textwidth}{!}{
    \begin{tabular}{l|cccc|c}
    \toprule
    & Imagenet-V2 & Imagenet-Sketch & Imagenet-A &  Imagenet-R  & Average\\
    \midrule
    CLIP \cite{radford2021learning} & 60.86 & 46.09 & 47.87 & 73.98 & 57.20 \\
    CLIP+TPT \cite{shu2022test} & 64.35 & 47.94 & 54.77 & 77.06 & 60.81 \\
    CoOp \cite{zhou2022coop} & {64.20} & 47.99  & 49.71  & 75.21  & {59.28} \\
    CoOp+TPT \cite{shu2022test} & \textbf{66.83} & 49.29  & 57.95  & 77.27  & 62.84 \\
    Co-CoOp \cite{zhou2022cocoop} & {64.07} & 48.75 & 50.63 & 76.18 & {59.91}  \\
    Co-CoOp+TPT \cite{shu2022test} & 64.85 & 48.27 & 58.47 & 78.65 & 62.61  \\
    Maple \cite{khattakMaPLe} & 64.07 & 49.15 & 50.90 & 76.98 & 60.28 \\
    Maple + TPT \cite{shu2022test} &64.87& 48.16& 58.08& 78.12 & 62.31 \\
    PromptAlign \cite{samadh2023align} & 65.29 & 50.23 & 59.37  & 79.33 & 63.55 \\
    DiffTPT \cite{feng2023diverse} & 65.10 & 46.80 & 55.68 &75.00 & 60.52  \\
    TDA \cite{karmanov2024efficient} & 64.67 & {50.54} & 60.11  & 80.24 & 63.89  \\
    \midrule
    \rowcolor{tabhighlight} \modelname &{65.51}	&\textbf{51.28}	&\textbf{64.53}	&\textbf{80.95}	&\textbf{65.57} \\
    \bottomrule
    \end{tabular}}
    \label{tab:dg}
\end{table}

\begin{table}[!t]
    \caption{\textnormal{\textbf{Full results on the Cross-Domain Benchmark with ViT-B/16 backbone.}} We report  top-1 accuracy and ``Average" is calculated by taking the mean accuracy across all ten datasets. The error bound is $\pm 0.17$.}
    \tabstyle{4pt}
    \resizebox{0.95\textwidth}{!}{
    \begin{tabular}{l|cccccccccc|c}
    \toprule
    & \rotatebox{90}{Caltech} & \rotatebox{90}{Pets} & \rotatebox{90}{Cars} & \rotatebox{90}{Flowers} & \rotatebox{90}{Food101} & \rotatebox{90}{Aircraft} & \rotatebox{90}{SUN397} & \rotatebox{90}{DTD} & \rotatebox{90}{EuroSAT} & \rotatebox{90}{UCF101} & \rotatebox{90}{\emph{ Average}} \\
    \midrule
    CLIP \cite{radford2021learning} & 93.35 & 88.25 & 65.48 & 67.44 & 83.65 & 23.67 & 62.59 & 44.27 & 42.01 & 65.13 & 63.58 \\
    CLIP+TPT \cite{shu2022test}  & 94.16 & 87.79 & 66.87 & 68.98 & 84.67 & 24.78 & 65.50 & \textbf{47.75} & 42.44 & 68.04 & 65.10 \\
    CoOp \cite{zhou2022coop}  & 93.70 & 89.14 & 64.51 & 68.71 & 85.30 & 18.47 & 64.15 & 41.92 & 46.39 & 66.55 & 63.88 \\
    CoCoOp \cite{zhou2022cocoop} & 93.79 &90.46  &64.90 &70.85  &83.97 &22.29 &66.89  & 45.45 & 39.23 & 68.44 & 64.63 \\
    MaPLe \cite{khattakMaPLe} & 93.53 & 90.49 & 65.57 & 72.23 & 86.20 & 24.74 & 67.01 & 46.49 & 48.06 & 68.69 & 66.30 \\
    MaPLe+TPT \cite{shu2022test} & 93.59 & 90.72 & 66.50 & 72.37 & 86.64 & 24.70 & {67.54} & 45.87 & 47.80 & 69.19 & 66.50 \\
    DiffTPT \cite{feng2023diverse} & 92.49 & 88.22 & 67.01 & 70.10  & 87.23 & 25.60 & 65.74 & 47.00 &43.13& 62.67 & 65.47 \\   
    PromptAlign \cite{samadh2023align} & 94.01 & \textbf{90.76} & 68.50 & \textbf{72.39} & {86.65} & 24.80 & {67.54} & 47.24 & 47.86 & 69.47 & 66.92 \\
    TDA \cite{karmanov2024efficient}& 94.24 & 88.63 & 67.28 & 71.42 & 86.14 & 23.91 & 67.62 & {47.40} & 58.00  & 70.66 & 67.53 \\
    \midrule
    \rowcolor{tabhighlight} \modelname &\textbf{94.77}	&89.51	&\textbf{69.30}	&{71.66}	&\textbf{87.17}	&\textbf{27.45}	&\textbf{68.09}	&{45.69}	&\textbf{61.22}	&\textbf{71.93}	&\textbf{68.68} \\
    \bottomrule
    \end{tabular}
    }
    \label{tab:cross}
\end{table}

\subsection{Out-of-Distribution Generalization}
To verify the robustness of BoostAdapter, we evaluate our method on the OOD benchmark, in comparison with existing training-require methods including CoOp \cite{zhou2022coop}, CoCoOp \cite{zhou2022cocoop}, TPT \cite{shu2022test}, DiffTPT \cite{feng2023diverse}, Maple \cite{khattakMaPLe} and PromptAlign \cite{samadh2023align}, as well as training-free method TDA \cite{karmanov2024efficient}. 
As can be seen from Table \ref{tab:dg}, the most striking observation emerging from 
the comparison is that \modelname{} significantly outperforms other baselines on average and improves the generalization ability of the model. 
For training-free methods such as TPT, DiffTPT and PromptAlign, BoostAdapter achieves superior performance while saving on optimization computation overhead.
For training-free methods like TDA, \modelname{} gains consistent performance improvements with the introduction of the boosting samples.
Notably, BoostAdapter surpasses TDA by 4.42\% on ImageNet-A and 0.84\% on ImageNet-V2, respectively.
This enhancement indicates the effectiveness of self-bootstrapping when historical samples may not provide sufficient useful information.

\begin{figure*}[!t]
	\centering
	\begin{subfigure}{0.32\linewidth}
		\includegraphics[width=\linewidth]{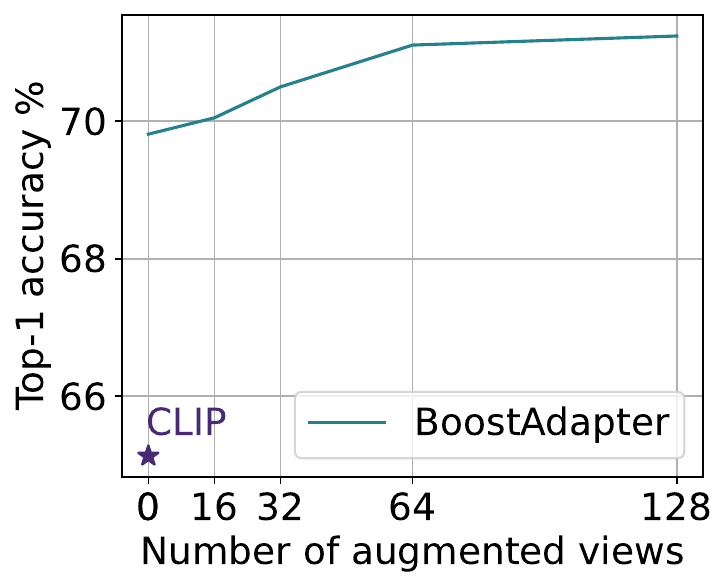}
		\caption{Number of augmented views}
		\label{fig:views}
	\end{subfigure}
	\hfill
	\begin{subfigure}{0.32\linewidth}
		\includegraphics[height=3.85cm, width=\linewidth]{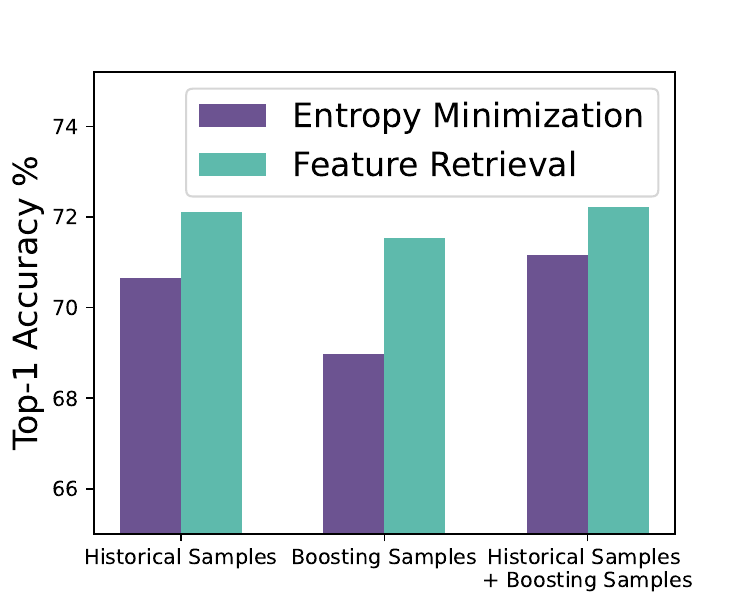}
		\caption{Adaptation methods}
		\label{fig:tpt}
	\end{subfigure}
	\hfill
	\begin{subfigure}{0.32\linewidth}
		\includegraphics[width=\linewidth]{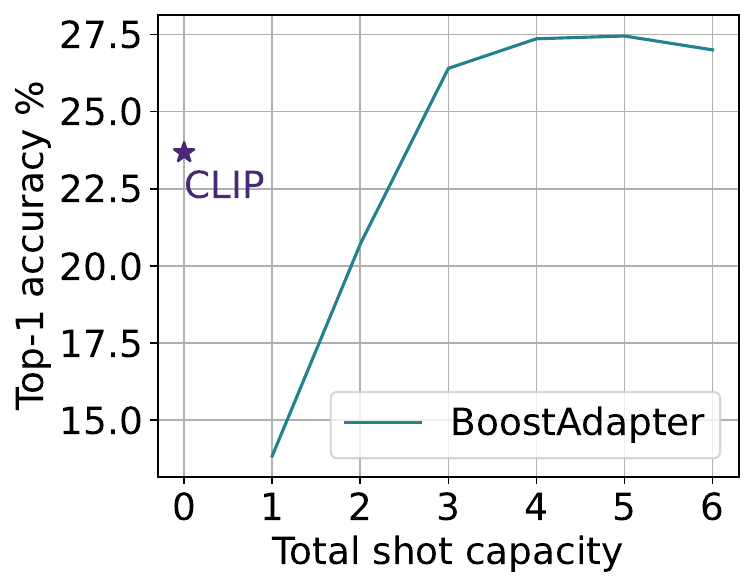}
		\caption{Total shot capacity}
		\label{fig:shots}
	\end{subfigure}
	\caption{Ablation studies of (a) number of augmented views to generate boosting samples (b) different adaptation methods and (c) total shot capacity of the cache. }
	\label{fig:analyses_all}
\end{figure*}
\begin{table*}[!t]
\begin{minipage}{0.5\textwidth}
    
    
    \captionof{table}{\textnormal{\textbf{Ablation study on historical samples and boosting samples on the OOD benchmark with ViT-B/16 backbone.} } We report top-1 accuracy and the error bound is $\pm 0.12$. }
    \small \centering
    \setlength{\tabcolsep}{8pt}
    \resizebox{\textwidth}{!}{
    \setlength{\tabcolsep}{3pt}
    \begin{tabular}{l|cccc|c}
    \toprule
    & -V2 & -Sketch & -A &  -R  & Average\\
    \midrule
    CLIP  & 60.86 & 46.09 & 47.87 & 73.98 & 57.20 \\
    Historical Samples  & 64.93 &50.23&63.80 & 80.43  & 64.85 \\ 
    Boosting Samples &65.40	&50.59	&64.40	&\textbf{80.96}	&65.34\\
    \midrule
    \rowcolor{tabhighlight} \modelname &\textbf{65.51}	&\textbf{51.28}	& \textbf{64.53}	&\textbf{80.95}	&\textbf{65.57} \\
    \bottomrule
    \end{tabular}}
    \label{tab:boosting}

\end{minipage}
\hspace{6pt}
\begin{minipage}{0.46\textwidth}
    
    \captionof{table}{\textnormal{\textbf{Full results on the OOD benchmark with RN-50 backbone.} } We report top-1 accuracy and the error bound is $\pm 0.06$. }
    \small \centering
 \setlength{\tabcolsep}{8pt}
    \resizebox{\textwidth}{!}{
    \begin{tabular}{l|cccc|c}
    \toprule
    & -V2 & -Sketch & -A &  -R  & Average\\
    \midrule
    CLIP \cite{radford2021learning} & 51.41	&33.37	&21.83	&56.15	&40.69\\
    TPT \cite{shu2022test} &54.70	&35.09	&26.67	&59.11	&43.89 \\
    CALIP \cite{guo2023calip} &53.70	&35.61	&23.96	&60.81	&43.52 \\
    CoOp \cite{zhou2022coop} &55.40	&34.67	&23.06	&56.60	&42.43\\
    CoCoOp \cite{zhou2022cocoop} &55.72	&34.48	&23.32	&57.74	&42.82\\
    DiffTPT \cite{feng2023diverse} &55.80	&37.10	&31.06	&58.80	&45.69\\
    TDA \cite{karmanov2024efficient} &55.54	& {38.12}	&30.29	&62.58	&46.63\\
    \midrule
    \rowcolor{tabhighlight} \modelname &\textbf{56.14}	&\textbf{38.87}	&\textbf{35.12}	&\textbf{62.66}	&\textbf{48.20} \\
    \bottomrule
    \end{tabular}}
    \label{tab:rn_dg}

\end{minipage}
\end{table*}

\subsection{Cross-Domain Transfer}
We further highlight our improvements in the transfer ability of CLIP on the Cross-Domain benchmark and present the results in Table \ref{tab:cross}. Compared with existing training-required and training-free methods, BoostAdapter achieves state-of-the-art performance on 7 out of 10 tasks, surpassing the strongest baselines by an average of 1.15\%. With diverse classes at test time, regional boosting enables BoostAdapter to adaptively extract knowledge that makes classes distinct from each other in a multi-scale manner. Notably, for datasets requiring fine-grained information for classification such as Aircraft, the improvement of BoostAdapter is most significant.

\begin{table}[!t]
    \caption{\textnormal{\textbf{Full results on the Cross-Domain Benchmark with RN-50 backbone.}} We report  top-1 accuracy and ``Average" is calculated by taking the mean accuracy across all ten datasets. The error bound is $\pm 0.05$.}
    \tabstyle{4pt}
    \resizebox{0.95\linewidth}{!}{
    \begin{tabular}{l|cccccccccc|c}
    \toprule
    & \rotatebox{90}{Caltech} & \rotatebox{90}{Pets} & \rotatebox{90}{Cars} & \rotatebox{90}{Flowers} & \rotatebox{90}{Food101} & \rotatebox{90}{Aircraft} & \rotatebox{90}{SUN397} & \rotatebox{90}{DTD} & \rotatebox{90}{EuroSAT} & \rotatebox{90}{UCF101} & \rotatebox{90}{\emph{ Average}} \\
    \midrule
    CLIP \cite{radford2021learning} &85.88	&83.57	&55.70	&61.75	&73.97	&15.66	&58.8	&40.37	&23.69	&58.84	&55.82\\
    CLIP + TPT \cite{shu2022test}&87.02	&84.49	&58.46	&62.69	&74.88	&17.58	&61.46	&40.84	&28.33	&60.82	&57.66\\
    CALIP \cite{guo2023calip}&87.71	&86.21	&56.27	&66.38	&77.42	&17.76	&58.59	&42.39	&38.90	&61.72	&59.34 \\
    DiffTPT \cite{feng2023diverse} &86.89	&83.40	&\textbf{60.71}	&63.53	&\textbf{79.21}	&17.60	&62.72	&40.72	&41.04	&62.67	&59.85\\
    CuPL \cite{pratt2023does} &89.29	&84.84	&57.28	&65.44	&76.94	&\textbf{19.59}	&62.55	&\textbf{48.64}	&38.38	&58.97	&60.19 \\
    TDA \cite{karmanov2024efficient} &\textbf{89.70}	&\textbf{86.18}	&57.78	&\textbf{68.74}	&77.75	&17.61	&62.53	&{43.74}	&\textbf{42.11}	&64.18	&61.03\\
    \midrule
    \rowcolor{tabhighlight} \modelname &{88.48}	&{85.75}	&{59.67}	&{68.25}	&{78.78}	&{18.93}	&\textbf{62.83}	&{43.85}	&\textbf{44.40}	&\textbf{64.42}	&\textbf{61.54} \\
    \bottomrule
    \end{tabular}
    }
    \label{tab:rn_cross}
\end{table}

\begin{table}[!t]
    \caption{\textbf{Comparisons with baselines on ImageNet-C at severity level 5 regarding accuracy (\%).}}
    \label{tab:imagnet-c}
 \begin{center}
 \LARGE
    \resizebox{1.0\linewidth}{!}{
 	\begin{tabular}{l|ccc|cccc|cccc|cccc|c}
     \toprule
 	\multicolumn{1}{c|}{} & \multicolumn{3}{c|}{Noise} & \multicolumn{4}{c|}{Blur} & \multicolumn{4}{c|}{Weather} & \multicolumn{4}{c|}{Digital}  \\
 	  & Gauss. & Shot & Impul. & Defoc. & Glass & Motion & Zoom & Snow & Frost & Fog & Brit. & Contr. & Elastic & Pixel & JPEG & Avg.  \\
    \cmidrule{1-17}
        CLIP-ViT-B/16 &15.15	&16.28	&15.26	&25.83	&16.87	&26.34	&24.43	&34.56	&33.01	&39.10 &57.78	&18.45	&14.71	&35.62	&35.81	&27.28 \\
        TDA &17.50	&18.59	&18.12	&59.12	&19.02	&28.25	&26.24	&37.30	&35.30	&41.57 &59.04	&21.06	&17.61	&37.78	&37.26	&31.58 \\
        \rowcolor{tabhighlight} \modelname &\textbf{17.53}	&\textbf{18.89}	&\textbf{18.39}	&\textbf{59.70}	&\textbf{19.07}	&\textbf{28.62}	&\textbf{27.33}	&\textbf{38.21}	&\textbf{36.13}	&\textbf{42.31} &\textbf{59.63}	&\textbf{21.22}	&\textbf{18.23}	&\textbf{39.25}	&\textbf{38.07}	&\textbf{32.17}   \\ 
        \bottomrule
	\end{tabular} 
        }
    \end{center}
\end{table}

\subsection{Ablation Study}
\label{sec:ablation}
\para{Historical Samples and Boosting Samples.} 
To demonstrate the effect of historical and boosting samples, we introduce two variants of BoostAdapter that utilize only historical samples or only boosting samples, respectively. Additionally, we provide the zero-shot results of CLIP for comparison. As shown in Table \ref{tab:boosting}, CLIP significantly benefits from both historical samples and boosting samples, resulting in notable improvements in performance. The consistent improvement of BoostAdapter compared to the variant that utilizes only historical samples further confirms the effectiveness of incorporating boosting samples into the training-free adapters. See Section \ref{sec:boosting_more} in the Appendix for more results.

\para{Number of Augmented Views for Boosting Samples.} 
We augment the testing samples and filter them by mutual information with the CLIP text embedding to obtain the boosting samples. We vary the number of augmented views and investigate the performance of BoostAdapter on UCF101 in Figure \ref{fig:views}. With a larger number of augmented views, the performance improves due to more bootstrapping information of the test sample, which is consistent with the conclusions of TPT \cite{shu2022test} and PromptAlign \cite{samadh2023align}. However, the computational overhead also increases with more augmented views, and selecting 64 augmented views is a fair trade-off between boosting performance and efficiency.

\para{Adaptation Methods.} 
Training-required methods use entropy as a self-supervised objective, whereas training-free methods classify samples based on feature retrieval. We compare the performance of these two adaptation methods under the constraints of historical samples only, boosting samples only, or both, and present the results on Flower102 in Fig. \ref{fig:tpt}.
Entropy minimization requires gradient descent and model optimization, resulting in high training costs and relatively lower performance across all three settings. In contrast, the training-free methods based on feature retrieval offer significant performance improvements with lower computational overhead. Additionally, both adaptation methods benefit from combining historical samples and boosting samples, consistent with the conclusions in Table \ref{tab:boosting}.

\para{Total shot capacity.}
BoostAdapter maintains low-entropy samples per class in the cache, and Figure \ref{fig:shots} studies the influence of different total shot capacities containing historical samples and boosting samples of each class on Aircraft. As can be observed from the results, when the cache capacity is small, the low-entropy samples maintained by BoostAdapter do not necessarily provide a benefit for classification compared to CLIP. As the shot capacity increases, BoostAdapter will achieve the best balance of diversity and relevance, and a larger capacity does not guarantee better performance.

\para{Versatility.} 
To demonstrate the versatility of BoostAdapter, we apply it to the RN-50 backbone and present the results in Tables \ref{tab:rn_dg} and \ref{tab:rn_cross}. The improvement is consistent and on average, BoostAdapter outperforms TDA by 1.57\% on the OOD benchmark and 0.49\% on the Cross-Domain benchmark.

\begin{table*}[!t]
\begin{minipage}{\textwidth}
    \captionof{table}{\textbf{Efficiency analysis.} We evaluate different methods on a single NVIDIA 3090 24GB GPU and report the frames per second (fps) and memory cost (GB).}
    \label{tab:time}
    \setlength{\tabcolsep}{8pt}
    \resizebox{\textwidth}{!}{
        \begin{tabular}{l|cccccc}
    \toprule
    & Augmentation  & Views & Inference Speed (fps) & Memory (GB) & OOD  Results	& Cross-Domain Results\\
    \midrule
    CLIP & - & - & 82.3 & 0.7 & 57.20	& 63.58\\
    TPT & Augmix & 64 &0.29 & 4.5 & 60.81	& 65.10\\
    DiffTPT & Diffusion & 64 & 0.10 & 14.4 & 	60.52	& 66.92\\
    TDA & Augmix & 64 & 11.89 & 1.2 & 	63.89 &	67.53  \\
    \midrule
    \rowcolor{tabhighlight} \modelname & Rand. Crop \& Rand. Horiz. Flip & 64 & 11.23 & 1.2 & 65.57& 68.68 \\
    \bottomrule
    \end{tabular}
    }
    \label{tab:efficiency}
    \centering

\end{minipage}
\end{table*}

\subsection{Discussions}
\para{Generalization on Corruption Datasets}
To further evaluate the generalization ability of BoostAdapter in new test-time scenarios, we compare BoostAdapter with baseline methods on the Imagenet-C dataset at the highest severity level 5. The key observation from Table \ref{tab:imagnet-c} is that BoostAdapter consistently outperforms TDA across all 15 corruption types, highlighting its practical applicability in real-world situations.
The superior performance of BoostAdapter stems from its capability to capture the knowledge of the test sample even under severe corruption. This is achieved with the help of the boosting samples, which effectively filter out noisy parts while retaining useful information.

\para{Efficiency Analysis}
BoostAdapter requires augmentation over the test samples, which may slightly affect the inference speed during testing. We conduct an efficiency analysis of BoostAdapter in comparison with existing Test Time Augmentation (TTA) methods and provide the results in Table \ref{tab:efficiency}. BoostAdapter is slightly slower than the cache-based method TDA, yet still significantly faster than training-required methods. The memory cost of BoostAdapter is also comparable to other baselines.

\begin{table*}[!t]
\begin{minipage}{0.48\textwidth}
    \captionof{table}{\textnormal{\textbf{Unification of more training-required methods.} BoostAdapter benefits from different training-required methods.}} 
    \small \centering
 \setlength{\tabcolsep}{8pt}
    \resizebox{\textwidth}{!}{
    \begin{tabular}{l|cccc|c}
    \toprule
    & -V & -S & -A &  -R  & Average\\
    \midrule
    CLIP-ViT-B/16 & 60.86 & 46.09 & 47.87 & 73.98 & 57.20  \\
    TDA & 64.67 & {50.54} & 60.11  & 80.24 & 63.89  \\
    \rowcolor{tabhighlight} \modelname &{65.51}	&{51.28 }	&{64.53}	&{80.95}	&{65.57} \\
    \rowcolor{tabhighlight} \modelname + TSD 	&{65.49}	&{51.50} &{64.37}	&{81.15}	&{65.63}  \\
    \rowcolor{tabhighlight} \modelname + DEYO 	&65.71	&51.52	&64.65	&81.43	&65.83  \\

    \bottomrule
    \end{tabular}}
    \label{tab:dg}
\end{minipage}
\hfill
\begin{minipage}{0.5\textwidth}
    \centering    \includegraphics[width=\textwidth]{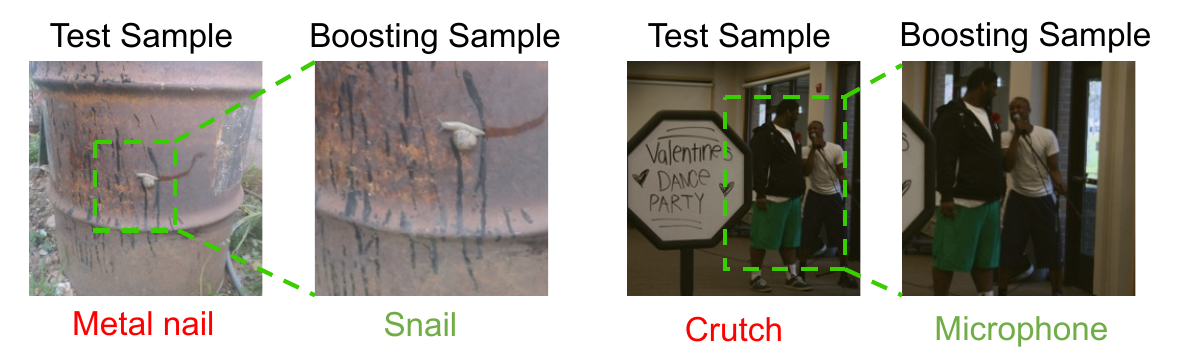}
    \captionof{figure}{\small  
    \textbf{Qualitative results.} The model predictions are provided below the images.  Boosting samples with low entropy improves information extraction from the test sample and helps the model to distinguish better. 
    }
    \label{fig:vis}
\end{minipage}
\end{table*}

\para{Unification of Training-required and Training-free Methods.}
From the unified perspective, we can also enhance training-free adapters with additional training-required methods. Here we take TSD \cite{wang2023feature} and DEYO \cite{lee2024entropy} as the showcase. Specifically, in the BoostAdapter+DEYO variant, we filter out augmented views with a PLPD lower than 0.2. For the BoostAdapter TSD variant, we discard augmented views that have different cache predictions and CLIP predictions to ensure consistency of the boosting samples. When equipping BoostAdapter with the technique of TSD and DEYO, we observe further improvement and find that training-free adapters can benefit from various boosting techniques of training-required methods.

\para{Qualitative Results}
The qualitative results are provided in Figure. \ref{fig:vis}. By incorporating samples with low entropy from regional bootstrapping, the model is enhanced to more effectively capture the fine-grained information of the test samples, thereby improving the overall performance.

\section{Conclusions}
\label{sec:conslusions}
In this work, we present an insightful analysis of existing training-required and training-free TTA methods to bridge the gap between them. In particular, we improve training-free adapters by incorporating self-boosting samples into the memory bank inspired by the idea of regional bootstrapping from entropy-based methods. The cache in our method, containing instance-agnostic historical samples and instance-aware boosting samples, is capable of performing knowledge mining on both the target domain and the testing sample itself. We also derive error bounds in the test-time adaptation setting and show that this cache benefits from both historical samples and boosting samples. Extensive experiments on the two benchmarks demonstrate the effectiveness of our method.

Despite the promising performance of our method, it also has some limitations. It requires slightly more computation overhead than existing training-free adapters due to the multiple augmentation of the test samples, as discussed in Appendix. One future direction is to develop a more efficient augmentation method to obtain boosting samples, rather than merely randomly cropping and then filtering over the test samples. 

\section*{Acknowledgements}
This work is supported in part by the National Natural Science Foundation of China, under Grant(62302309,62171248),  Shenzhen Science and Technology Program (JCYJ20220818101014030, JCYJ20220818101012025), and the PCNL KEY project (PCL2023AS6-1).

\bibliographystyle{plainnat}
\bibliography{main}

\begin{thebibliography}{56}
\providecommand{\natexlab}[1]{#1}
\providecommand{\url}[1]{\texttt{#1}}
\expandafter\ifx\csname urlstyle\endcsname\relax
  \providecommand{\doi}[1]{doi: #1}\else
  \providecommand{\doi}{doi: \begingroup \urlstyle{rm}\Url}\fi

\bibitem[Bommasani et~al.(2021)Bommasani, Hudson, Adeli, Altman, Arora, von Arx, Bernstein, Bohg, Bosselut, Brunskill, et~al.]{bommasani2021opportunities}
Rishi Bommasani, Drew~A Hudson, Ehsan Adeli, Russ Altman, Simran Arora, Sydney von Arx, Michael~S Bernstein, Jeannette Bohg, Antoine Bosselut, Emma Brunskill, et~al.
\newblock On the opportunities and risks of foundation models.
\newblock \emph{arXiv preprint arXiv:2108.07258}, 2021.

\bibitem[Bossard et~al.(2014)Bossard, Guillaumin, and Van~Gool]{bossard2014food}
Lukas Bossard, Matthieu Guillaumin, and Luc Van~Gool.
\newblock Food-101--mining discriminative components with random forests.
\newblock In \emph{Computer Vision--ECCV 2014: 13th European Conference, Zurich, Switzerland, September 6-12, 2014, Proceedings, Part VI 13}, pages 446--461. Springer, 2014.

\bibitem[Cimpoi et~al.(2014)Cimpoi, Maji, Kokkinos, Mohamed, and Vedaldi]{cimpoi2014describing}
Mircea Cimpoi, Subhransu Maji, Iasonas Kokkinos, Sammy Mohamed, and Andrea Vedaldi.
\newblock Describing textures in the wild.
\newblock In \emph{Proceedings of the IEEE Conference on Computer Vision and Pattern Recognition}, pages 3606--3613, 2014.

\bibitem[Deng et~al.(2009)Deng, Dong, Socher, Li, Li, and Fei-Fei]{deng2009imagenet}
Jia Deng, Wei Dong, Richard Socher, Li-Jia Li, Kai Li, and Li~Fei-Fei.
\newblock Imagenet: A large-scale hierarchical image database.
\newblock In \emph{2009 IEEE Conference on Computer Vision and Pattern Recognition}, pages 248--255. IEEE, 2009.

\bibitem[Fei-Fei et~al.(2004)Fei-Fei, Fergus, and Perona]{fei2004learning}
Li~Fei-Fei, Rob Fergus, and Pietro Perona.
\newblock Learning generative visual models from few training examples: An incremental bayesian approach tested on 101 object categories.
\newblock In \emph{2004 Conference on Computer Vision and Pattern Recognition Workshop}, pages 178--178. IEEE, 2004.

\bibitem[Feng et~al.(2023)Feng, Yu, Liu, Khan, and Zuo]{feng2023diverse}
Chun-Mei Feng, Kai Yu, Yong Liu, Salman Khan, and Wangmeng Zuo.
\newblock Diverse data augmentation with diffusions for effective test-time prompt tuning.
\newblock In \emph{Proceedings of the IEEE/CVF International Conference on Computer Vision}, pages 2704--2714, 2023.

\bibitem[Gao et~al.(2024{\natexlab{a}})Gao, Gu, Bai, Xia, Torr, Liu, and Li]{gao2024energy}
Kuofeng Gao, Jindong Gu, Yang Bai, Shu-Tao Xia, Philip Torr, Wei Liu, and Zhifeng Li.
\newblock Energy-latency manipulation of multi-modal large language models via verbose samples.
\newblock \emph{arXiv preprint arXiv:2404.16557}, 2024{\natexlab{a}}.

\bibitem[Gao et~al.(2024{\natexlab{b}})Gao, Geng, Zhang, Ma, Fang, Zhang, Li, and Qiao]{gao2024clip}
Peng Gao, Shijie Geng, Renrui Zhang, Teli Ma, Rongyao Fang, Yongfeng Zhang, Hongsheng Li, and Yu~Qiao.
\newblock Clip-adapter: Better vision-language models with feature adapters.
\newblock \emph{International Journal of Computer Vision}, 132\penalty0 (2):\penalty0 581--595, 2024{\natexlab{b}}.

\bibitem[Guo et~al.(2024)Guo, Dai, Ouyang, Zhang, Zha, Chen, and Xia]{guo2024refir}
Hang Guo, Tao Dai, Zhihao Ouyang, Taolin Zhang, Yaohua Zha, Bin Chen, and Shu-tao Xia.
\newblock Refir: Grounding large restoration models with retrieval augmentation.
\newblock \emph{arXiv preprint arXiv:2410.05601}, 2024.

\bibitem[Guo et~al.(2023)Guo, Zhang, Qiu, Ma, Miao, He, and Cui]{guo2023calip}
Ziyu Guo, Renrui Zhang, Longtian Qiu, Xianzheng Ma, Xupeng Miao, Xuming He, and Bin Cui.
\newblock Calip: Zero-shot enhancement of clip with parameter-free attention.
\newblock In \emph{Proceedings of the AAAI Conference on Artificial Intelligence}, volume~37, pages 746--754, 2023.

\bibitem[Helber et~al.(2019)Helber, Bischke, Dengel, and Borth]{helber2019eurosat}
Patrick Helber, Benjamin Bischke, Andreas Dengel, and Damian Borth.
\newblock Eurosat: A novel dataset and deep learning benchmark for land use and land cover classification.
\newblock \emph{IEEE Journal of Selected Topics in Applied Earth Observations and Remote Sensing}, 12\penalty0 (7):\penalty0 2217--2226, 2019.

\bibitem[Hendrycks and Dietterich(2019)]{hendrycks2019benchmarking}
Dan Hendrycks and Thomas Dietterich.
\newblock Benchmarking neural network robustness to common corruptions and perturbations.
\newblock \emph{arXiv preprint arXiv:1903.12261}, 2019.

\bibitem[Hendrycks et~al.(2021{\natexlab{a}})Hendrycks, Basart, Mu, Kadavath, Wang, Dorundo, Desai, Zhu, Parajuli, Guo, et~al.]{hendrycks2021many}
Dan Hendrycks, Steven Basart, Norman Mu, Saurav Kadavath, Frank Wang, Evan Dorundo, Rahul Desai, Tyler Zhu, Samyak Parajuli, Mike Guo, et~al.
\newblock The many faces of robustness: A critical analysis of out-of-distribution generalization.
\newblock In \emph{Proceedings of the IEEE/CVF International Conference on Computer Vision}, pages 8340--8349, 2021{\natexlab{a}}.

\bibitem[Hendrycks et~al.(2021{\natexlab{b}})Hendrycks, Zhao, Basart, Steinhardt, and Song]{hendrycks2021natural}
Dan Hendrycks, Kevin Zhao, Steven Basart, Jacob Steinhardt, and Dawn Song.
\newblock Natural adversarial examples.
\newblock In \emph{Proceedings of the IEEE/CVF Conference on Computer Vision and Pattern Recognition}, pages 15262--15271, 2021{\natexlab{b}}.

\bibitem[Iwasawa and Matsuo(2021)]{iwasawa2021test}
Yusuke Iwasawa and Yutaka Matsuo.
\newblock Test-time classifier adjustment module for model-agnostic domain generalization.
\newblock \emph{Advances in Neural Information Processing Systems}, 34:\penalty0 2427--2440, 2021.

\bibitem[Jia et~al.(2021)Jia, Yang, Xia, Chen, Parekh, Pham, Le, Sung, Li, and Duerig]{jia2021scaling}
Chao Jia, Yinfei Yang, Ye~Xia, Yi-Ting Chen, Zarana Parekh, Hieu Pham, Quoc Le, Yun-Hsuan Sung, Zhen Li, and Tom Duerig.
\newblock Scaling up visual and vision-language representation learning with noisy text supervision.
\newblock In \emph{ICML}, pages 4904--4916. PMLR, 2021.

\bibitem[Karmanov et~al.(2024)Karmanov, Guan, Lu, Saddik, and Xing]{karmanov2024efficient}
Adilbek Karmanov, Dayan Guan, Shijian Lu, Abdulmotaleb~El Saddik, and Eric Xing.
\newblock Efficient test-time adaptation of vision-language models.
\newblock \emph{arXiv preprint arXiv:2403.18293}, 2024.

\bibitem[Khattak et~al.(2023)Khattak, Rasheed, Maaz, Khan, and Khan]{khattakMaPLe}
Muhammad~Uzair Khattak, Hanoona Rasheed, Muhammad Maaz, Salman Khan, and Fahad~Shahbaz Khan.
\newblock Maple: Multi-modal prompt learning.
\newblock In \emph{The IEEE/CVF Conference on Computer Vision and Pattern Recognition}, 2023.

\bibitem[Krause et~al.(2013)Krause, Stark, Deng, and Fei-Fei]{krause20133d}
Jonathan Krause, Michael Stark, Jia Deng, and Li~Fei-Fei.
\newblock 3d object representations for fine-grained categorization.
\newblock In \emph{Proceedings of the IEEE International Conference on Computer Vision Workshops}, pages 554--561, 2013.

\bibitem[Kumari et~al.(2023)Kumari, Zhang, Zhang, Shechtman, and Zhu]{kumari2023multi}
Nupur Kumari, Bingliang Zhang, Richard Zhang, Eli Shechtman, and Jun-Yan Zhu.
\newblock Multi-concept customization of text-to-image diffusion.
\newblock In \emph{Proceedings of the IEEE/CVF Conference on Computer Vision and Pattern Recognition}, pages 1931--1941, 2023.

\bibitem[Lee et~al.(2024)Lee, Jung, Lee, Park, Shin, Hwang, and Yoon]{lee2024entropy}
Jonghyun Lee, Dahuin Jung, Saehyung Lee, Junsung Park, Juhyeon Shin, Uiwon Hwang, and Sungroh Yoon.
\newblock Entropy is not enough for test-time adaptation: From the perspective of disentangled factors.
\newblock \emph{arXiv preprint arXiv:2403.07366}, 2024.

\bibitem[Lester et~al.(2021)Lester, Al-Rfou, and Constant]{lester2021power}
Brian Lester, Rami Al-Rfou, and Noah Constant.
\newblock The power of scale for parameter-efficient prompt tuning.
\newblock \emph{arXiv preprint arXiv:2104.08691}, 2021.

\bibitem[Li et~al.(2021)Li, Selvaraju, Gotmare, Joty, Xiong, and Hoi]{li2021align}
Junnan Li, Ramprasaath Selvaraju, Akhilesh Gotmare, Shafiq Joty, Caiming Xiong, and Steven Chu~Hong Hoi.
\newblock Align before fuse: Vision and language representation learning with momentum distillation.
\newblock \emph{Advances in neural information processing systems}, 34:\penalty0 9694--9705, 2021.

\bibitem[Li et~al.(2022)Li, Li, Xiong, and Hoi]{li2022blip}
Junnan Li, Dongxu Li, Caiming Xiong, and Steven Hoi.
\newblock Blip: Bootstrapping language-image pre-training for unified vision-language understanding and generation.
\newblock In \emph{International conference on machine learning}, pages 12888--12900. PMLR, 2022.

\bibitem[Li et~al.(2023)Li, Li, Savarese, and Hoi]{li2023blip}
Junnan Li, Dongxu Li, Silvio Savarese, and Steven Hoi.
\newblock Blip-2: Bootstrapping language-image pre-training with frozen image encoders and large language models.
\newblock In \emph{International conference on machine learning}, pages 19730--19742. PMLR, 2023.

\bibitem[Li et~al.(2024)Li, Lian, Lu, Bai, Chen, and Wang]{li2024graphadapter}
Xin Li, Dongze Lian, Zhihe Lu, Jiawang Bai, Zhibo Chen, and Xinchao Wang.
\newblock Graphadapter: Tuning vision-language models with dual knowledge graph.
\newblock \emph{Advances in Neural Information Processing Systems}, 36, 2024.

\bibitem[Liang et~al.(2017)Liang, Li, and Srikant]{liang2017enhancing}
Shiyu Liang, Yixuan Li, and Rayadurgam Srikant.
\newblock Enhancing the reliability of out-of-distribution image detection in neural networks.
\newblock \emph{arXiv preprint arXiv:1706.02690}, 2017.

\bibitem[Liu et~al.(2021)Liu, Ji, Fu, Tam, Du, Yang, and Tang]{liu2021p}
Xiao Liu, Kaixuan Ji, Yicheng Fu, Weng~Lam Tam, Zhengxiao Du, Zhilin Yang, and Jie Tang.
\newblock P-tuning v2: Prompt tuning can be comparable to fine-tuning universally across scales and tasks.
\newblock \emph{arXiv preprint arXiv:2110.07602}, 2021.

\bibitem[Lu et~al.(2022)Lu, Liu, Zhang, Liu, and Tian]{lu2022prompt}
Yuning Lu, Jianzhuang Liu, Yonggang Zhang, Yajing Liu, and Xinmei Tian.
\newblock Prompt distribution learning.
\newblock In \emph{Proceedings of the IEEE/CVF Conference on Computer Vision and Pattern Recognition}, pages 5206--5215, 2022.

\bibitem[Lu et~al.(2023)Lu, Bai, Li, Xiao, and Wang]{lu2023beyond}
Zhihe Lu, Jiawang Bai, Xin Li, Zeyu Xiao, and Xinchao Wang.
\newblock Beyond sole strength: Customized ensembles for generalized vision-language models.
\newblock \emph{arXiv preprint arXiv:2311.17091}, 2023.

\bibitem[Maji et~al.(2013)Maji, Rahtu, Kannala, Blaschko, and Vedaldi]{maji2013fine}
Subhransu Maji, Esa Rahtu, Juho Kannala, Matthew Blaschko, and Andrea Vedaldi.
\newblock Fine-grained visual classification of aircraft.
\newblock \emph{arXiv preprint arXiv:1306.5151}, 2013.

\bibitem[Nilsback and Zisserman(2008)]{nilsback2008automated}
Maria-Elena Nilsback and Andrew Zisserman.
\newblock Automated flower classification over a large number of classes.
\newblock In \emph{2008 Sixth Indian Conference on Computer Vision, Graphics \& Image Processing}, pages 722--729. IEEE, 2008.

\bibitem[Niu et~al.(2023)Niu, Wu, Zhang, Wen, Chen, Zhao, and Tan]{niu2023towards}
Shuaicheng Niu, Jiaxiang Wu, Yifan Zhang, Zhiquan Wen, Yaofo Chen, Peilin Zhao, and Mingkui Tan.
\newblock Towards stable test-time adaptation in dynamic wild world.
\newblock \emph{arXiv preprint arXiv:2302.12400}, 2023.

\bibitem[Parkhi et~al.(2012)Parkhi, Vedaldi, Zisserman, and Jawahar]{parkhi2012cats}
Omkar~M Parkhi, Andrea Vedaldi, Andrew Zisserman, and CV~Jawahar.
\newblock Cats and dogs.
\newblock In \emph{2012 IEEE Conference on Computer Vision and Pattern Recognition}, pages 3498--3505. IEEE, 2012.

\bibitem[Pratt et~al.(2023)Pratt, Covert, Liu, and Farhadi]{pratt2023does}
Sarah Pratt, Ian Covert, Rosanne Liu, and Ali Farhadi.
\newblock What does a platypus look like? generating customized prompts for zero-shot image classification.
\newblock In \emph{Proceedings of the IEEE/CVF International Conference on Computer Vision}, pages 15691--15701, 2023.

\bibitem[Radford et~al.(2021)Radford, Kim, Hallacy, Ramesh, Goh, Agarwal, Sastry, Askell, Mishkin, Clark, et~al.]{radford2021learning}
Alec Radford, Jong~Wook Kim, Chris Hallacy, Aditya Ramesh, Gabriel Goh, Sandhini Agarwal, Girish Sastry, Amanda Askell, Pamela Mishkin, Jack Clark, et~al.
\newblock Learning transferable visual models from natural language supervision.
\newblock In \emph{International Conference on Machine Learning}, pages 8748--8763. PMLR, 2021.

\bibitem[Recht et~al.(2019)Recht, Roelofs, Schmidt, and Shankar]{recht2019imagenet}
Benjamin Recht, Rebecca Roelofs, Ludwig Schmidt, and Vaishaal Shankar.
\newblock Do imagenet classifiers generalize to imagenet?
\newblock In \emph{International Conference on Machine Learning}, pages 5389--5400. PMLR, 2019.

\bibitem[Rombach et~al.(2022)Rombach, Blattmann, Lorenz, Esser, and Ommer]{rombach2022high}
Robin Rombach, Andreas Blattmann, Dominik Lorenz, Patrick Esser, and Bj{\"o}rn Ommer.
\newblock High-resolution image synthesis with latent diffusion models.
\newblock In \emph{Proceedings of the IEEE/CVF conference on computer vision and pattern recognition}, pages 10684--10695, 2022.

\bibitem[Samadh et~al.(2023)Samadh, Gani, Hussein, Khattak, Naseer, Khan, and Khan]{samadh2023align}
Jameel Hassan~Abdul Samadh, Hanan Gani, Noor~Hazim Hussein, Muhammad~Uzair Khattak, Muzammal Naseer, Fahad Khan, and Salman Khan.
\newblock Align your prompts: Test-time prompting with distribution alignment for zero-shot generalization.
\newblock In \emph{Thirty-seventh Conference on Neural Information Processing Systems}, 2023.

\bibitem[Shen et~al.(2018)Shen, Qu, Zhang, and Yu]{shen2018wasserstein}
Jian Shen, Yanru Qu, Weinan Zhang, and Yong Yu.
\newblock Wasserstein distance guided representation learning for domain adaptation.
\newblock In \emph{Proceedings of the AAAI conference on artificial intelligence}, volume~32, 2018.

\bibitem[Shu et~al.(2022)Shu, Nie, Huang, Yu, Goldstein, Anandkumar, and Xiao]{shu2022test}
Manli Shu, Weili Nie, De-An Huang, Zhiding Yu, Tom Goldstein, Anima Anandkumar, and Chaowei Xiao.
\newblock Test-time prompt tuning for zero-shot generalization in vision-language models.
\newblock \emph{Advances in Neural Information Processing Systems}, 35:\penalty0 14274--14289, 2022.

\bibitem[Soomro et~al.(2012)Soomro, Zamir, and Shah]{soomro2012dataset}
Khurram Soomro, Amir~Roshan Zamir, and Mubarak Shah.
\newblock A dataset of 101 human action classes from videos in the wild.
\newblock \emph{Center for Research in Computer Vision}, 2\penalty0 (11), 2012.

\bibitem[Wang et~al.(2020)Wang, Shelhamer, Liu, Olshausen, and Darrell]{wang2020tent}
Dequan Wang, Evan Shelhamer, Shaoteng Liu, Bruno Olshausen, and Trevor Darrell.
\newblock Tent: Fully test-time adaptation by entropy minimization.
\newblock \emph{arXiv preprint arXiv:2006.10726}, 2020.

\bibitem[Wang et~al.(2019)Wang, Ge, Lipton, and Xing]{wang2019learning}
Haohan Wang, Songwei Ge, Zachary Lipton, and Eric~P Xing.
\newblock Learning robust global representations by penalizing local predictive power.
\newblock \emph{Advances in Neural Information Processing Systems}, 32, 2019.

\bibitem[Wang et~al.(2023{\natexlab{a}})Wang, Zhang, Yan, Zhang, and Li]{wang2023feature}
Shuai Wang, Daoan Zhang, Zipei Yan, Jianguo Zhang, and Rui Li.
\newblock Feature alignment and uniformity for test time adaptation.
\newblock In \emph{Proceedings of the IEEE/CVF Conference on Computer Vision and Pattern Recognition}, pages 20050--20060, 2023{\natexlab{a}}.

\bibitem[Wang et~al.(2023{\natexlab{b}})Wang, Zhang, Cen, Gao, Zhang, Zhao, and Sang]{wang2023clip}
Xiang Wang, Shiwei Zhang, Jun Cen, Changxin Gao, Yingya Zhang, Deli Zhao, and Nong Sang.
\newblock Clip-guided prototype modulating for few-shot action recognition.
\newblock \emph{International Journal of Computer Vision}, pages 1--14, 2023{\natexlab{b}}.

\bibitem[Wasim et~al.(2023)Wasim, Naseer, Khan, Khan, and Shah]{wasim2023vita}
Syed~Talal Wasim, Muzammal Naseer, Salman Khan, Fahad~Shahbaz Khan, and Mubarak Shah.
\newblock Vita-clip: Video and text adaptive clip via multimodal prompting.
\newblock In \emph{Proceedings of the IEEE/CVF Conference on Computer Vision and Pattern Recognition}, pages 23034--23044, 2023.

\bibitem[Xiao et~al.(2010)Xiao, Hays, Ehinger, Oliva, and Torralba]{xiao2010sun}
Jianxiong Xiao, James Hays, Krista~A Ehinger, Aude Oliva, and Antonio Torralba.
\newblock Sun database: Large-scale scene recognition from abbey to zoo.
\newblock In \emph{2010 IEEE Computer Society Conference on Computer Vision and Pattern Recognition}, pages 3485--3492. IEEE, 2010.

\bibitem[Yang et~al.(2022)Yang, Duan, Tran, Xu, Chanda, Chen, Zeng, Chilimbi, and Huang]{yang2022vision}
Jinyu Yang, Jiali Duan, Son Tran, Yi~Xu, Sampath Chanda, Liqun Chen, Belinda Zeng, Trishul Chilimbi, and Junzhou Huang.
\newblock Vision-language pre-training with triple contrastive learning.
\newblock In \emph{Proceedings of the IEEE/CVF Conference on Computer Vision and Pattern Recognition}, pages 15671--15680, 2022.

\bibitem[Yang et~al.(2024)Yang, Bai, Gao, Yang, Li, and Xia]{yang2024not}
Sheng Yang, Jiawang Bai, Kuofeng Gao, Yong Yang, Yiming Li, and Shu-Tao Xia.
\newblock Not all prompts are secure: A switchable backdoor attack against pre-trained vision transfomers.
\newblock In \emph{CVPR}, 2024.

\bibitem[Zha et~al.(2023)Zha, Wang, Dai, Chen, Wang, and Xia]{zha2023instance}
Yaohua Zha, Jinpeng Wang, Tao Dai, Bin Chen, Zhi Wang, and Shu-Tao Xia.
\newblock Instance-aware dynamic prompt tuning for pre-trained point cloud models.
\newblock In \emph{Proceedings of the IEEE/CVF International Conference on Computer Vision}, pages 14161--14170, 2023.

\bibitem[Zhang et~al.(2022)Zhang, Zhang, Fang, Gao, Li, Dai, Qiao, and Li]{zhang2022tip}
Renrui Zhang, Wei Zhang, Rongyao Fang, Peng Gao, Kunchang Li, Jifeng Dai, Yu~Qiao, and Hongsheng Li.
\newblock Tip-adapter: Training-free adaption of clip for few-shot classification.
\newblock In \emph{European conference on computer vision}, pages 493--510. Springer, 2022.

\bibitem[Zhang et~al.(2023)Zhang, Wang, Jin, Yuan, Zhang, Wang, Jin, and Tan]{zhang2023adanpc}
Yifan Zhang, Xue Wang, Kexin Jin, Kun Yuan, Zhang Zhang, Liang Wang, Rong Jin, and Tieniu Tan.
\newblock Adanpc: Exploring non-parametric classifier for test-time adaptation.
\newblock In \emph{International Conference on Machine Learning}, pages 41647--41676. PMLR, 2023.

\bibitem[Zhou et~al.(2022{\natexlab{a}})Zhou, Yang, Loy, and Liu]{zhou2022cocoop}
Kaiyang Zhou, Jingkang Yang, Chen~Change Loy, and Ziwei Liu.
\newblock Conditional prompt learning for vision-language models.
\newblock In \emph{IEEE/CVF Conference on Computer Vision and Pattern Recognition (CVPR)}, 2022{\natexlab{a}}.

\bibitem[Zhou et~al.(2022{\natexlab{b}})Zhou, Yang, Loy, and Liu]{zhou2022coop}
Kaiyang Zhou, Jingkang Yang, Chen~Change Loy, and Ziwei Liu.
\newblock Learning to prompt for vision-language models.
\newblock \emph{International Journal of Computer Vision (IJCV)}, 2022{\natexlab{b}}.

\bibitem[Zhu et~al.(2023)Zhu, Zhang, He, Zhou, Wang, Zhao, and Gao]{zhu2023not}
Xiangyang Zhu, Renrui Zhang, Bowei He, Aojun Zhou, Dong Wang, Bin Zhao, and Peng Gao.
\newblock Not all features matter: Enhancing few-shot clip with adaptive prior refinement.
\newblock In \emph{Proceedings of the IEEE/CVF International Conference on Computer Vision}, pages 2605--2615, 2023.

\end{thebibliography}

\clearpage
\appendix
\section*{\Large{Appendix}}
\label{sec:Appendix}

\section{Dataset and Licenses}
\label{sec:licenses}
Table \ref{tab:datasets} presents the statistics and details of datasets used in the paper. We also provide the corresponding license information of the datasets and source code. 

\para{Datasets.}  
Below are the datasets used in this paper that have known license information:
The following datasets used in this paper are under the MIT License: ImageNet-A~\citep{hendrycks2021natural}, ImageNet-V2~\citep{recht2019imagenet}, ImageNet-R~\citep{hendrycks2021many}, ImageNet-Sketch~\citep{wang2019learning}, EuroSAT \cite{helber2019eurosat}, Food101 \cite{bossard2014food}. \\
The following datasets used in this paper are under the CC BY-SA 4.0 License: Oxford-Pets~\citep{parkhi2012cats}, Caltech101 \cite{fei2004learning}. \\
The following datasets used in this paper are for research purposes only: DTD~\citep{cimpoi2014describing}, StanfordCars~\citep{krause20133d}, SUN397~\citep{xiao2010sun}, FGVC-Aircraft~\citep{maji2013fine}, Flower102~\citep{parkhi2012cats}, UCF101 \cite{soomro2012dataset}. 

\para{Source code.}
We use the implementation of existing baseline methods for reporting their results in this paper. Below are their license information:
Source code used in this paper that are under the MIT License: {CLIP}~\citep{radford2021learning}, {PromptAlign}~\citep{samadh2023align} and TDA~\citep{karmanov2024efficient}.

\begin{table}[h]
    \centering
    \renewcommand\arraystretch{1}{
    \begin{tabular}{l c c c}
    \toprule
    Dataset & Description & Classes & Test Size\\
    \midrule
    \multicolumn{4}{c}{Out-of-Distribution Benchmark}\\
    \midrule
    ImageNet-V2 & New Validation Sets of ImageNet  &1,000  & 10,000\\
    ImageNet-S  & Sketch Images &1,000 & 
    50,000\\
    ImageNet-A & Natural Adversarial Examples &200  & 7,500 \\
    ImageNet-R & Rendition Extension of ImageNet &200 & 30,000 \\
    \midrule
    \multicolumn{4}{c}{Cross-Domain Benchmark}\\
    \midrule
    Aircraft & Aircraft Model Classification &100 & 3,333 \\
    Caltech101 & Natural Image Classification &100 & 2,465 \\
    Cars & Cars Classification &196 & 8,041 \\
    DTD &Describable Textures Dataset &47 & 1,692 \\
    EuroSAT & Satellite Images &10 & 8,100 \\
    Flowers102 &Flowers Classification& 102 & 2,463 \\
    Food101 &Food Classification &101 & 30,300 \\
    Pets &Pets Classification &37 & 3,669 \\
    SUN397 & Scene Categorization Benchmark &397 & 19,850 \\
    UCF101 & Action Recognition Dataset &101 & 3,783 \\
    \bottomrule
    \end{tabular}}
    \vspace{10pt}
    \caption{Datasets statistics.}
    \label{tab:datasets}
\end{table}

\section{Broader Impacts}
\label{sec:impacts}
In this paper, we focus on bridging the gap between training-required and training-free methods to improve the generalization ability of vision-language models. We also theoretically derive the error bound of incorporating boosting samples into the historical cache. We hope that our work will inspire the community to explore test-time adaptation in an effective and efficient way.

\section{Theoretical Proof}
\label{sec:proof}
\subsection{Cross-entropy Optimization behaves like Cache Classifier over well-clustered Samples (Proof of Proposition \ref{prop:eq})}
Given well-clustered samples in the feature space and the classifier defined in \myref{equ:clip}, we first derive the distance between the weights of the classifier and the optimal weights and then establish the connection between the optimal weights with the features center of the samples.

Suppose the classifier function $f$ over samples is convex and differentiable, and also $L$-smooth. Let the distance between initial weight $w^{(0)}$ and optimal weight $w^*$ to be $D=||w^{(0)}-w^*||$. GD updates by ${w}^{(t+1)}={w}^t-f^*_t\nabla f({w}^t)$ with step size $f^*_t=\frac1L$, and then GD enjoys the following convergence guarantee:
\begin{equation}
||w-w^*||\leq\frac{2L\left\|w^{(0)}-{w}^\star\right\|^2}{T-1}=\mathcal{O}\left(\frac{LD^2}{T}\right).
\label{equ:converge}
\end{equation}

We then showcase the relationship between $w_*$ and the features center $\mu_i$ of class $i, i=1,2,...,N$.
Since we optimize on well-clustered samples, we consider the scenarios of perfect clusters, where samples in the class $i$ will be encoded into the same point $\mu_i$ by the encoder $g$, and these points should be farthest enough between each other. 
Given $n$ samples $\{(x_k,y_k)\}_{k=1}^n$, with the number of samples in class $i$ to be $n_i$, the cross-entropy loss function $L$ can be written as:
\begin{equation}
L=-\sum_{i=1}^n\log P(y=y_k|x_k)
\end{equation}
Substitute the sample $g(x_k)=\mu_i$ from class $i$, we derive the probability $P(y=i|x_k)$ using the softmax function from \myref{equ:clip} is:
\begin{equation}
P(y=i|x_k)=\frac{exp({w_i^T\mu_i})}{\sum_{j=1}^Nexp({w_j^T\mu_i})} .
\end{equation}
Thus, the cross-entropy loss for a sample $(x_k,y_k=i)$ is:
\begin{equation}
L_k=-\log\left(\frac{exp({w_{i}^T\mu_{i}})}{\sum_{j=1}^Nexp({w_j^T\mu_{i}})}\right) .
\end{equation}

For all samples, the total loss is:
\begin{equation}
L=-\sum_{i=1}^N n_i \log\left(\frac{exp({w_{i}^T\mu_{i}})}{\sum_{j=1}^Nexp({w_j^T\mu_i})}\right) .\end{equation}

The gradient of the loss with respect to $w_i$ can be simplified as:
\begin{equation}
\frac{\partial L}{\partial w_i}=-\mu_in_i+\mu_i\sum_{k=1}^Nn_k\frac{\exp(w_i^T\mu_k)}{\sum_{j=1}^N\exp(w_j^T\mu_k)} .
\end{equation}

When converges to the optimal weight, we have the condition of fixed point $\frac{\partial L}{\partial w_i^*}=0$. And we have 
\begin{equation}
    -\mu_in_i+\mu_i\sum_{k=1}^Nn_k\frac{\exp((w_i^*)^T\mu_k)}{\sum_{j=1}^N\exp((w_j^*)^T\mu_k)}=0 .
\end{equation}

Thus, we have 
\begin{equation}
    \sum_{k=1}^Nn_k\frac{\exp((w_i^*)^T\mu_k)}{\sum_{j=1}^N\exp((w_j^*)^T\mu_k)} = n_i .
    \label{equ:eq}
\end{equation}

Given a well-clustered samples, we could have $exp((w^*_i )^T \mu_k) \gg exp((w^*_j )^T \mu_k)$ for a specific $i$ when $w^*_i$ is near $\mu_k$. Then since 
the equality in \myref{equ:eq} will hold for each class and for class $i=1,2,...,N$ we have 
\begin{equation}
    w_i^* \to \mu_i .
    \label{equ:optimal}
\end{equation}

Combining \myref{equ:converge} and \myref{equ:eq}, with iteration steps $T$, we show that the weight of classfier will finally converge to the feature center of each class:
\begin{equation}
    ||w-\mu||\leq||w-w^*||+||w^*-\mu||\le\mathcal{O}\left(\frac{LD^2}{T}\right).
    \label{equ:w_mu}
\end{equation}
And we have the output logits of the optimal weights with the encoder $g$:
\begin{equation}
    \boldsymbol{p}_{cross}(x) = [\mu_1^T g(x), \mu_2^T g(x), ..., \mu_N^T g(x)]
    \label{equ:pred_cross}
\end{equation}

Next we discess the behavior of the cache classifier over these samples. 
Given the number of well-clustered samples in class $i$ to be $n_i$, the output logits of the cache classifier defined in \myref{equ:cache_simple} using samples $\{(x_k,y_k)\}_{k=1}^n$ can be described as follows:
\begin{align}
    \boldsymbol{p}_{cache}(x)&=\sum_{k=1}^n \frac{1}{n_{y_i}} [g(x_k)^T g(x)] y_k\notag \\
    &=\sum_{i=1}^N \frac{n_i}{n_i}[\mu_i^T g(x)] y_i\notag\\
    &= [\mu_1^T g(x), \mu_2^T g(x), ..., \mu_N^T g(x)]
    \label{eq:pred_knn}
\end{align}
Combining \myref{equ:pred_cross} and \myref{eq:pred_knn}, we draw the conclusion that cross-entropy optimization behaves like cache classifier over well-clustered samples. 

\subsection{Historical Cache reduce Empirical Risk (Proof of Proposition \ref{prop:knn_his})}\label{sec:theo}
We follow the proofs in \cite{zhang2023adanpc} and extend the conclusion to boosting samples. 
\subsubsection{Additional Definitions and Assumptions}
\begin{definition}
(\textbf{Wasserstein-distance and the dual form}). Wasserstein distance measures the distance between two probability distributions on a given metric space. It is defined using the concept of optimal transport. For two distributions $\mathbb{P}, \mathbb{Q}$, The $\rho$-th Wasserstein distance is defined as
\begin{equation}
W_p(\mathbb{P},\mathbb{Q})=\left(\inf_{\gamma\in\Pi(\mathbb{P},\mathbb{Q})}\int_{X\times X}d(x,y)^pd\gamma(x,y)\right)^{1/p}
\label{eq:wd}
\end{equation}
Here, $\Pi(\mathbb{P},\mathbb{Q})$ denotes the set of all couplings (or transport plans) $\gamma$ of $\mathbb{P}$ and $\mathbb{Q}$,  \ie joint
distributions on $X\times X$ with marginals $\mathbb{P}$ and $\mathbb{Q}$.The idea is to find the optimal way to transport the
mass from one distribution to the other with the minimal cost, where the cost is given by the $p$-th power of the distance.

The first Wasserstein distance, $W_{1}(\mathbb{P},\mathbb{Q})$ ,often referred to as the Earth-Mover Distance(EMD), has a particularly elegant dual representation. 
The dual form of $W_{1}$ leverages the Kantorovich-Rubinstein duality and can be expressed as:
\begin{equation}
W_1(\mathbb{P},\mathbb{Q})=\sup_{\|f\|_{\mathrm{Lip}}\leq1}\left(\int_Xf\:d\mathbb{P}-\int_Xf\:d\mathbb{Q}\right)
\end{equation}
Here, the supremum is taken over all 1-Lipschitz functions $f$ ,which are functions satisfying $|f(x)-$ $f(y)|\leq d(x,y)$ for all $x,y\in X$ .This representation shows that $W_{1}$ can be seen as the maximum difference in expected values of a 1-Lipschitz function over the two distributions. In the following part, Wasserstein distance represents the first Wasserstein distance for simplicity and we utilize $W(\cdot, \cdot)$ instead of $W_1(\cdot, \cdot)$.
\end{definition}

Given the definition of the Wasserstein distance, we have the following proposition that derive the empirical risk on the target domain according to Theorem 1 from \cite{shen2018wasserstein}. 

\begin{prop}
Given two distributions $\mathbb{P},\mathbb{Q}$, denote $f^*=\arg\min_{f\in\mathcal{H}}(\epsilon_P(f)+\epsilon_Q(f))$ and $\xi =\epsilon_P(f^*)+\epsilon_Q(f^*)$. Assume all hypotheses $h$ are $L$-Lipschitz continuous, the risk of hypothesis $\hat{f}$ is then bounded by
\begin{equation}
    \epsilon_Q(\hat{f})\leq \xi+\epsilon_P(\hat{f})+2L\mathcal{W}(\mathbb{P},\mathbb{Q}).
    \label{lemma:bound}
\end{equation}
\label{prop:wass}
\end{prop}

\subsubsection{Distance between the Ball Distribution with the Target Distribution}
When using the cache classifier with historical samples, a large number of samples that are not similar enough from the target domain will be filtered and the selected samples with high weight are all close to the target data. Thus we extend the conclusion in \cite{zhang2023adanpc} to the distance between the ball distribution with the target distribution.
Considering a test sample from the target distribution ${x_t}\in p_t(x)$ and a distribution consisting of ball center of all the test samples $\Omega:=\bigcup_{x_t\in p_t(x)} \mathcal{B}({x_t},r)$, 
informally, according to \myref{eq:wd}, we have the distance between the ball distribution with the target distribution as follows:
\begin{equation}
\mathcal{W}(\Omega,p_t(x))=\inf_{\gamma\in \Pi[\Omega,p_t(x)]} \iint \parallel {x_{t}}-{x_{ball}}\parallel  d\gamma({x_{t}},{x_{ball}}),
\end{equation}
where for each $x_{ball}\in\Omega$, we can find at least one $x_t\in p_t(x)$ such that $\parallel x_{ball}-x_t\parallel\leq r$, the overall distance will then be bounded by $r$. 
Specifically, we can choose a density function $\gamma^*$ where $\gamma^*(x_{ball},x_t)>0$ only if $\parallel x_{ball}-x_t\parallel\leq r$ otherwise 0, then we have

\begin{align}
\mathcal{W}(\Omega,p_t(x))&=\inf_{\gamma\in \Pi[\Omega,p_t(x)]} \iint \parallel {x_{ball}}-{x_t}\parallel  d\gamma({x_{ball}},{x_t})\notag \\ 
&\leq \iint \parallel {x_{ball}}-{x_t}\parallel \gamma^*({x_{ball}},{x_t})  d x_{ball}x_t\leq r .
\end{align}
However, there is no guarantee that each data $x_t\in p_t(x)$ can find a neighbor $\mathcal{B}(x_t,r)$ with $|\mathcal{B}(x_t,r)|>0$ with all the small $r$. We then provide the probability that the set of neighbors $\mathcal{B}(x_t,r)$ of each $x_t\in p_t(x)$ is not measuring zero with respect to the radius $r$.  

As defined in the cache classfier \myref{equ:cache_simple}, we denote $k_t$ is the number of historical samples we select in the cache and $n_t$ is the total number of data from the historical stream. With the strong density assumption, given the coefficient bound $m$ and $M$,  for any $x_t \in p_t(x), r<R$, according to Assumption~\ref{assump1}, we have 

\begin{align}
|\hat{x_t}\in p_t(x) \wedge \hat{x_t} \in \mathcal{B}(x_t,r)|&=\int_{\mathcal{B}(x_t,r)\cap p_t(x)} \frac{d p_t(x)}{d\lambda}(\hat{x_t})d\hat{x_t}\notag\\
&\geq  m\lambda(\mathcal{B}(x_t,r)\cap p_t(x))\notag\\
&\geq mc_t\pi_dr^d,
\label{equ:a}
\end{align}

where $\pi_d=\lambda(\mathcal{B}(0,1))$ is the volume of the $d$ dimension unit ball and $\lambda$ is the Lebesgue measure of a set in a Euclidean space. Set $r_0=(\frac{2k}{mc_t \pi_dn_t})^{1/d}$, with a additional assumption that we utilize a small $k_t$ compared to $n_t$ so that $\frac{k_t}{n_t}<\frac{c_t m\pi_dr_\mu^d}{2}$, we have $r_0<R$. Then for any $x_t \in p_t(x)$, according to \myref{equ:a}, we have
\begin{equation}
    |\hat{x_t}\in p_t(x) \wedge \hat{x_t} \in \mathcal{B}(x_t,r_0)| \geq  mc_t\pi_dr_0^d >\frac{2k_t}{n_t} .
\end{equation}

Since $\hat{x_t}\in p_t(x)$ are independently drawn from the target distribution, let $\mathbb{I}(\cdot)$ to be the Indicator funciton and $S_{n_t}(x_t)=\sum_{i=1}^{n_t}\mathbb{I}(\hat{x_t}\in \mathcal{B}(x_t,r_0))$ denote the number of data $\hat{x_t}\in p_t(x) $ that fall into $\mathcal{B}(x_t,r_0)$, then $S_{n_t}(x_t)$ follows the Binomial distribution. Let $W\sim Binomial(n_t,\frac{2k}{n_t})$, according to the Chernoff inequality, we have 

\begin{align}
    P(S_{n_t}(x_t)<k_t)&\leq P(W<k_t)\notag\\
    &= P(W-\mathbb{E}[W]<-k_t)\notag\\
    &\leq \exp(-k_t^2/2\mathbb{E}[W])\notag\\
    &=\exp(-k_t/4),
\end{align}

where the second inequality holds since $S_n(x)$ has a larger mean than $W$. 
With a large $k_t$, the probability that $S_n(x)<k_t$ is small for any $x_t \in p_t(x)$. 
Denoting $\hat{x_t}^{(i)}$ as the $i_{th}$ nearest sample to $x_t$ among $\mathcal{B}(x_t,r_0)$  in the cache, we have for any $x_t \in p_t(x)$

\begin{equation}
P(\parallel \hat{x_t}^{(k_t)}-x_t   \parallel\leq r_0)= P(S_n(x_t)\geq k_t)\geq 1-\exp(-k_t/4)
\label{bound:x_diff}
\end{equation}
Combine \myref{bound:x_diff} with the assumption that the distribution $p_t(x)$ is finite with cardinality $\aleph_{p_t}$ and the desired probability part is shown by union bound.  

\begin{align}
\bigcap_{x_t \in p_t(x)} P(\parallel \hat{x_t}^{(k_t)}-x_t   \parallel\leq r_0))&=\bigcap_{x_t \in p_t(x)} P(S_n(x)\geq k_t)\notag\\
&=1-\bigcup_{x_t \in p_t(x)}P(S_n(x)< k_t)\notag\\
&\geq 1-\aleph_{p_t}\exp\left(-\frac{k_t}{4}\right)\notag\\
&= 1-\exp\left(-\frac{k_t}{4}+\log \aleph_{p_t}\right).
\end{align}

And then we have the following proposition.
\begin{prop}
Given the target domain distributions $p_t(x)$ that is finite with cardinality $\aleph_{p_t}$, and $\Omega:=\bigcup_{x\in p_t(x)} \mathcal{B}(x,r)$, where $\mathcal{B}(x,r)=\{x':\parallel x'-x\parallel\leq r\}$ denotes a ball centered on $x$ with radius $r$. Denote $f^*=\arg\min_{f\in\mathcal{H}}(\epsilon_t(f)+\epsilon_\Omega(f))$ and $\xi =\epsilon_t(f^*)+\epsilon_\Omega(f^*)$. Assume all hypotheses $h$ are $L$-Lipschitz continuous, the risk of hypothesis $\hat{f}$ on the unseen target domain is then bounded by
\begin{equation}
\begin{aligned}
\epsilon_t(\hat{f})\leq \kappa+\epsilon_\Omega(\hat{f})+2L\left(\frac{2k_t}{mc_t\pi_dn_t}\right)^{1/d}.   
\end{aligned}
\end{equation}
with probability $1-\exp(-\frac{k_t}{4}+\log \aleph_{p_t}) $
\label{prop:wass_omega}
\end{prop}

\subsubsection{Excess Error Bound of Cache Classifier}
Let $s_i$ to be the softmax probability $softmax(\boldsymbol{p}_{cache})$ for class $i$ in the the cache classifier from \myref{equ:cache_simple}, we can simplify the classifier as $\hat{f}_{cache}=\mathbb{I} \{ s_1 \geq \frac{1}{2}\}$ on the binary classification setting. Then $\hat{f}_{cache}(x_t)\neq f^*(x_t)$ implies that $\left| \hat{f}_{cache}(x_t) -f^*(x_t)\right|\geq \left|f^*(x_t)-\frac{1}{2}\right|$. We then bridge the gap between the excess error and the classify error as follows:
\begin{equation}
    \mathcal{E}_t(\hat{f})=2\mathbb{E}_{x_t\sim p_t(x)}\left[\left|f^*(x_t)-\frac{1}{2}\right|\mathbb{I}\left\{\left| \hat{f}_{cache}(x_t) -f^*(x_t)\right|\geq \left|f^*(x_t)-\frac{1}{2}\right|\right\}\right] .
\end{equation}

We want to bound $\sup_{x_t} \left| \hat{f}_{cache}(x_t) -f^*(x_t) \right|\leq t$, combining with the marginal assumption in Assumption~\ref{define_noise} and the fact that 
\begin{equation}
    \mathbb{E}\left[Z\cdot \mathbb{I}\{Z\leq t\}\right]\leq tP(Z\leq t),
\label{equ:indcator}
\end{equation}
where $Z=\left|f^*(x_t)-\frac{1}{2}\right|$, so
we have $ \mathcal{E}_t(\hat{f})\leq C_\beta t^{\beta+1}$. To bound $\left| \hat{f}_{cache}(x_t) -f^*(x_t) \right|$, we denote $(\hat{x_t}^{(i)},\hat{y_t}^{(i)})$ as the $i_{th}$ nearest data and the corresponding labels to $x_t$ in $\mathcal{B}(x_t,r_0)$. 
The result of the cache classfier with normalized weight will be 
\begin{align}
\hat{f}_{cache}(x_t)&=\sum_{i=1}^{k_t}\frac{1}{\sum_{j=1}^{k_t}\left[g\left(\hat{x_t}^{(j)}\right)^T g(x)\right]} \left[g\left(\hat{x_t}^{(i)}\right)^T g(x)\right]\hat{y_t}^{(i)}\\
&=\sum_{i=1}^{k_t}w_i \hat{y_t}^{(i)}, 
\end{align}
where $w_i = \frac{g\left(\hat{x_t}^{(i)}\right)^T g(x)}{\sum_{j=1}^{k_t}\left[g\left(\hat{x_t}^{(j)}\right)^T g(x)\right]}$ is the normalized weight and $\sum_{i=1}^{k_t} w_i=1$. 
Based on the assumptions and notions above, we have for any $x_t \in p_t(x)$

\begin{equation}
\begin{aligned}
\left|\hat{f}_{cache}(x_t) -f^*(x_t) \right |&=\left| \sum_{i=1}^{k_t} w_i \hat{y_t}^{(i)}-f^*(x_t)  \right|\\
& \leq \left|\sum_{i=1}^{k_t} w_i \hat{y_t}^{(i)}-\sum_{i=1}^{k_t} w_if^*\left(\hat{x_t}^{(i)}\right)\right|+\left| \sum_{i=1}^{k_t} w_if^*\left(\hat{x_t}^{(i)}\right)-f^*(x_t)  \right|\\
&\leq \underbrace{\left| \sum_{i=1}^{k_t} w_i \hat{y_t}^{(i)}-\sum_{i=1}^{k_t} f^*\left(\hat{x_t}^{(i)}\right)\right|}_{\rm \circled{\rm 1}}+\underbrace{\sum_{i=1}^{k_t} w_i\left| f^*\left(\hat{x_t}^{(i)}\right)-f^*(x_t)  \right|}_{\rm \circled{\rm 2}},
\end{aligned}
\label{equ:bound_cova}
\end{equation}
where ${\rm \circled{\rm 2}}$ is easy to bound. According to the assumption that $f^*$ is $C$-Smoothness, we have
\begin{equation}
\sum_{i=1}^{k_t} w_i\left| f^*\left(\hat{x_t}^{(i)}\right)-f^*(x_t)  \right|\leq  \sum_{i=1}^{k_t} {w_i} C \cdot \parallel \hat{x_t}^{(i)}- x_t \parallel \leq C \cdot \parallel \hat{x_t}^{(k_t)}- x_t \parallel
\label{equ:19}
\end{equation}
According to \myref{bound:x_diff}, with probability at least $1-\exp(-k_t/4)$, ${\rm \circled{\rm 2}}\leq C\left(\frac{2k_t}{mc_t\pi_dn_t}\right)^{1/d}$. 
Note that We store the target sample into the cache only when its prediction confidence is large enough. Therefore, it is natural to assume that:
\begin{equation}
    E_{Y|X}\left[\hat{y_t}^{(i)}\right]=f^*(x_t^{(i)}).
    \label{equ:mean_t}
\end{equation}
Then we use the Hoeffding inequality to obtain the upper bound of ${\rm \circled{\rm 1}}$
\begin{align}
P_{X,Y}&\left(\left|\sum_{i=1}^{k_t} w_i \hat{y_t}^{(i)}-\sum_{i=1}^{k_t}f^*(\hat{x_t}^{(i)})\right|>\epsilon\right) \notag\\
&=\mathbb{E}_X\left[P_{Y|X}\left(\left|\sum_{i=1}^{k_t}  w_i\hat{y_t}^{(i)}-\sum_{i=1}^{k_t}f^*(\hat{x_t}^{(i)})\right|>\epsilon\right)\right]\notag\\
&\leq 2\exp(-\frac{2\epsilon^2}{\sum_{i=1}^{k_t}w_i^2})\notag\\
&\approx 2\exp(-2\eta k_t\epsilon^2).
\label{equ:21}
\end{align}
We simplify the bound by assuming that the weights in the target domain are evenly distributed in the subset of all samples with respect to a specific class controlled by coefficient $\eta$, according to Assumption \ref{assump1} and Proposition \ref{prop:wass}. That is, we have $\sum_{i=1}^{k_t}w_i^2\approx \sum_{i=1}^{\eta k_t}{ {(\frac{1}{\eta k_t}})}^2 = \frac{1}{\eta k_t}$.

Set $\epsilon=(1/k_t)^{1/4}$, we have, with probability, at least $1-3\exp(-2\eta\sqrt{k_t})$, ${\rm \circled{\rm 1}}\leq(1/k_t)^{1/4}$, ${\rm \circled{\rm 2}}\leq C\left(\frac{2k_t}{mc_t\pi_dn_t}\right)^{1/d}$, and then $\left|\hat{f}_{cache}(x_t) -f^*(x_t) \right |\leq (1/k_t)^{1/4}+C\left(\frac{2k_t}{mc_t\pi_dn_t}\right)^{1/d}$. According to \myref{bound:x_diff} and \myref{equ:indcator}, the excess error is bounded by

\begin{align}
\mathcal{E}_t(\hat{f})&\leq 2C_\beta\left( \left(\frac{1}{k_t}\right)^{1/4}+  C\left(\frac{2k_t}{mc_t\pi_dn_t}\right)^{1/{d}}\right)^{{1+\beta}} \notag\\
&\approx \left(\left(\frac{1}{k_t}\right)^{1/4}+C_1  \left(\frac{k_t}{c_t n_t}\right)^{1/{d}} \right)^{{1+\beta}},
\label{bound:22}
\end{align}
with constant $C_1$. 
When appropriately choosing $k_t = \mathcal{O}(\log n_t)$,  we have 
\begin{equation}
\begin{aligned}
&\min\{1-2\exp(-2\eta\sqrt{k_t}),1-\exp(-k_t/4)\}\\ 
&\geq 1-2\exp(-2\eta\sqrt{k_t})-\exp(-k_t/4)\\
&\geq 1-3\exp(-2\eta\sqrt{k_t})\\
&=1-3\exp(-\mathcal{O}(1)\sqrt{\log n_t})
\end{aligned}
\end{equation}
where the third line is because $k_t/4>2\eta\sqrt{k_t}$ for large enough $k_t$. Namely, with probability at least $ 1-3\exp(-\sqrt{\log n_t})^{\mathcal{O}(1)}$, the following bound holds true.

\begin{equation}
\mathcal{E}_t(\hat{f})\leq  \mathcal{O}\left(\left(\frac{1}{\log n_t}\right)^{1/4}+\left(\frac{\log n_t}{c_t n_t}\right)^{1/{d}} \right)^{{1+\beta}},\label{eq:xue_2022_nov_11_1}
\end{equation}

\subsection{Historical Cache benefits from Boosting Samples (Proof of Proposition \ref{prop:knn_his_boost})}\label{sec:app_target_sample}
To study the effect of the boosting samples, we consider the cache classfier containing both $k_t$ historical samples $\{\hat{x_t}^{(i)},\hat{y_t}^{(i)}\}_{i=1}^{k_t}$ and $k_b$ boosting samples $\{\hat{x_b}^{(i)},\hat{y_b}^{(i)}\}_{i=1}^{k_b}$ as the nearest data to $x_t$ in $\mathcal{B}(x_t,r_0)$. 
With the normalized weights $w_{ti}=\frac{g\left(\hat{x_t}^{(i)}\right)^T g(x)}{\sum_{j=1}^{k_t}\left[g\left(\hat{x_t}^{(j)}\right)^T g(x)\right] + \sum_{j=1}^{k_b}\left[g\left(\hat{x_b}^{(j)}\right)^T g(x)\right]}$ and $w_{bi}=\frac{g\left(\hat{x_b}^{(i)}\right)^T g(x)}{\sum_{j=1}^{k_t}\left[g\left(\hat{x_t}^{(j)}\right)^T g(x)\right] + \sum_{j=1}^{k_b}\left[g\left(\hat{x_b}^{(j)}\right)^T g(x)\right]}$, 
the prediction result of the cache classifier will be $\hat{f}_{cache}(x_t)=\sum_{i=1}^{k_t} w_{ti} \hat{y_t}^{(i)}+\sum_{i=1}^{k_b} w_{bi}y_b^{(i)}$. 
Then we have:

\begin{equation}
\begin{aligned}
&\left|\hat{f}_{cache}(x_t) -f^*(x_t) \right | \notag\\
&=\left| \sum_{i=1}^{k_t} w_{ti}  \hat{y_t}^{(i)}-\sum_{i=1}^{k_t} w_{ti} f^*(x_t) +\sum_{i=1}^{k_b} w_{bi}  y_u^{(i)}-\sum_{i=1}^{k_b} w_{bi} f^*(x_t)  \right| \notag\\
& \leq \Bigg| \left[\sum_{i=1}^{k_t} w_{ti}  \hat{y_t}^{(i)}-\sum_{i=1}^{k_t} w_{ti}  f^*(\hat{x_t}^{(i)})\right]+\left[ \sum_{i=1}^{k_t} w_{ti}  f^*(\hat{x_t}^{(i)})-\sum_{i=1}^{k_t} w_{ti}f^*(x_t)  \right] \\
&\quad +\left[\sum_{i=1}^{k_b} w_{bi}  y_u^{(i)}-\sum_{i=1}^{k_b} w_{bi}  f^*\left(x_u^{(i)}\right)\right]+\left[ \sum_{i=1}^{k_b} w_{bi}  f^*\left(x_u^{(i)}\right)-\sum_{i=1}^{k_b} w_{bi}f^*(x_t)  \right]\Bigg| \\
&\leq \underbrace{\left|\sum_{i=1}^{k_t} w_{ti} \hat{y_t}^{(i)}+\sum_{i=1}^{k_b} w_{bi} y_u^{(i)}-\sum_{i=1}^{k_t} w_{ti} f^*(\hat{x_t}^{(i)})-\sum_{i=1}^{k_b} w_{bi} f^*\left(x_u^{(i)}\right)\right|}_{\rm \circled{\rm 1}}\\
&\quad+\underbrace{\sum_{i=1}^{k_t} w_{ti} \left| f^*(\hat{x_t}^{(i)})-f^*(x_t)  \right|}_{\rm \circled{\rm 2}} +\underbrace{\sum_{i=1}^{k_b} w_{bi} \left| f^*\left(x_u^{(i)}\right)-f^*(x_t)  \right|}_{\rm \circled{\rm 3}}
\end{aligned}
\end{equation}

Similar to \myref{equ:mean_t}, we have the following assumption on the boosting distribution:
\begin{equation}
    E_{Y|X}\left[\hat{y_b}^{(i)}\right]=f^*(x_b^{(i)}).
    \label{equ:mean_b}
\end{equation}

According to \myref{equ:21}, we have

\begin{equation}
\begin{aligned}
&P_{X,Y}\left(\left|\sum_{i=1}^{k_t} w_{ti} \hat{y_t}^{(i)}+\sum_{i=1}^{k_b} w_{bi} y_b^{(i)}-\sum_{i=1}^{k_t} w_{ti} f^*(\hat{x_t}^{(i)})-\sum_{i=1}^{k_b}w_{bi}f^*\left(x_b^{(i)}\right)\right|\right)\\
&=\mathbb{E}_X\left[P_{Y|X}\left(\left|\sum_{i=1}^{k_t} w_{ti} \hat{y_t}^{(i)}+\sum_{i=1}^{k_b} w_{bi} y_b^{(i)}-\sum_{i=1}^{k_t} w_{ti} f^*(\hat{x_t}^{(i)})-\sum_{i=1}^{k_b}w_{bi}f^*\left(x_b^{(i)}\right)\right|\right)\right]\\
&\leq 2\exp({-2\eta(k_t+k_b)\epsilon^2})
\end{aligned}
\label{equ:27}
\end{equation}

Set $\epsilon=(1/(k_t+k_b))^{1/4}$, we have, with probability, at least $1-3\exp(-2\eta\sqrt{(k_t+k_b)})$, ${\rm \circled{\rm 1}}\leq (1/(k_t+k_b))^{1/4}$. Then, according to \myref{equ:19}, we have
\begin{equation}
    \sum_{i=1}^{k_t} w_{ti}\left| f^*\left(\hat{x_t}^{(i)}\right)-f^*(x_t)  \right|\leq  \sum_{i=1}^{k_t} {w_{ti}} C \cdot \parallel \hat{x_t}^{(i)}- x_t \parallel \leq S_t C \cdot \parallel \hat{x_t}^{(k_t)}- x_t \parallel
\end{equation}
and
\begin{equation}
    \sum_{i=1}^{k_b} w_{bi}\left| f^*\left(\hat{x_b}^{(i)}\right)-f^*(x_t)  \right|\leq  \sum_{i=1}^{k_b} {w_{bi}} C  \cdot \parallel \hat{x_b}^{(i)}- x_t \parallel \leq S_b C  \cdot \parallel \hat{x_b}^{(k_b)}- x_t \parallel .
\end{equation}
where $S_t=\sum_{i=1}^{k_t} {w_{ti}}$, $S_b=\sum_{i=1}^{k_b} {w_{bi}}$ are the sum of weights of historical samples and boosting samples, respectively, and we have $S_t + S_b = 1$.

Then we have the following results in similar:
\begin{equation}
{\rm \circled{\rm 2}}\leq S_t C\left(\frac{2k_t}{mc_t\pi_dn_t}\right)^{1/d}; {\rm \circled{\rm 3}}\leq S_b C\left(\frac{2k_b}{mc_b\pi_dn_b}\right)^{1/d}
\end{equation}
Finally, the excess error under the covariate shift setting can be bounded by
\begin{equation}
\begin{aligned}
\mathcal{E}_t(\hat{f})&\leq 2C_\beta\left( 
(\frac{1}{k_t+k_b})^{1/4}
+ S_t C\left(\frac{2k_t}{mc_t\pi_dn_t}\right)^{1/d} + S_b C\left(\frac{2k_b}{mc_b\pi_dn_b}\right)^{1/d}
\right)^{{1+\beta}}\\
& \approx  \left(\left(\frac{1}{k_t+k_b}\right)^{1/4}+C_1 S_t  \left(\frac{k_t}{c_t n_t}\right)^{1/{d}} +C_1 S_b \left(\frac{k_b}{c_b n_b}\right)^{1/{d}}  \right)^{{1+\beta}} \\
& =  \left(\left(\frac{1}{k_t+k_b}\right)^{1/4}+C_1 \sum_{i=1}^{k_t} {w_{ti}}  \left(\frac{k_t}{c_t n_t}\right)^{1/{d}} +C_1 \sum_{i=1}^{k_b} {w_{bi}} \left(\frac{k_b}{c_b n_b}\right)^{1/{d}}  \right)^{{1+\beta}} 
\end{aligned}
\label{bound:28}
\end{equation}
Compared \myref{bound:28} to \myref{bound:22} and $S_t + S_b =1$, it is easy to verify that
\begin{equation}
\begin{aligned}
& {(S_t+S_b)}C\left(\frac{2(k_t+k_b)}{mc_t\pi_dn_t}\right)^{1/{d}}-S_t C\left(\frac{2k_t}{mc_t\pi_dn_t}\right)^{1/d} - S_b C\left(\frac{2k_b}{mc_b\pi_dn_b}\right)^{1/d}\\
&\geq S_b C\left(\frac{2k_t}{mc_t\pi_dn_t}\right)^{1/d}- S_b C\left(\frac{2k_b}{mc_b\pi_dn_b}\right)^{1/d} 
\label{equ:diff}
\end{aligned}
\end{equation}
In general, the boosting distribution is more close to the test sample than the target distribution and we have $c_b>c_t$.
Thus the difference in \myref{equ:diff} is then larger than $0$, namely incorporating boosting samples into the memory bank, the excess error can be further reduced.

\section{More Experiments}
\para{Independent Cache for Boosting Samples.}
In BoostAdater, due to the cost of augmentation, the number of boosting samples is relatively smaller than the number of historical samples. Therefore, we use a joint cache for storing both historical and boosting samples to facilitate intra-sample and inter-sample interactions. Table \ref{tab:ind_cache_dg} and Table \ref{tab:ind_cache_cross} study the influence of using an independent cache for the boosting samples. As can be observed from the results, BoostAdapter suffers from slight performance degradation due to the independent cache.
\begin{table}[!h]
    \caption{\textnormal{\textbf{Independent cache for boosting samples on the OOD benchmark.} }} 
    \small \centering
 \setlength{\tabcolsep}{8pt}
    \resizebox{\textwidth}{!}{
    \begin{tabular}{l|cccc|c}
    \toprule
    & Imagenet-V2 & Imagenet-Sketch & Imagenet-A &  Imagenet-R  & Average\\
    \midrule
    Independent Cache &65.37	&50.62	&\textbf{64.56}	&\textbf{80.96}	&65.38\\
    \midrule
    \rowcolor{tabhighlight} Joint Cache &\textbf{65.51}	&\textbf{51.28}	&{64.53}	&\textbf{80.95}	&\textbf{65.57} \\
    \bottomrule
    \end{tabular}}
    \label{tab:ind_cache_dg}
\end{table}

\begin{table}[h]
    \caption{\textnormal{\textbf{Independent cache for boosting sample on the Cross-Domain Benchmark.}}}
    \tabstyle{4pt}
    \resizebox{\textwidth}{!}{
        \begin{tabular}{l|cccccccccc|c}
            \toprule
                                                           & \rotatebox{90}{Caltech} & \rotatebox{90}{Pets} & \rotatebox{90}{Cars} & \rotatebox{90}{Flowers} & \rotatebox{90}{Food101} & \rotatebox{90}{Aircraft} & \rotatebox{90}{SUN397} & \rotatebox{90}{DTD} & \rotatebox{90}{EuroSAT} & \rotatebox{90}{UCF101} & \rotatebox{90}{\emph{ Average}} \\
            \midrule
            Independent Cache                              &94.69	&88.88	&69.19	&\textbf{71.94}	&86.99	&26.76	&67.64	&44.21	&61.20	&69.63	&68.11                           \\
            \rowcolor{tabhighlight} Joint Cache &\textbf{94.77}	&\textbf{89.51}	&\textbf{69.30}	&{71.66}	&\textbf{87.17}	&\textbf{27.45}	&\textbf{68.09}	&\textbf{45.69}	&\textbf{61.22}	&\textbf{71.93}	&\textbf{68.68}                      \\
            \bottomrule
        \end{tabular}
    }
    \label{tab:ind_cache_cross}
\end{table}

\para{Different Augmentation for Boosting Samples.}
We make use of random crop followed by random horizontal flip as augmentations for generating boosting samples. Additionally, we further explore the influences of different augmentations applied to the randomly cropped images. The comparison methods include: 
(i) Random Brighness: Randomly set the brighness of image from 50\% to 150\%. 
(ii) Random Auto Contrast: Apply auto contrast over image with probability $p=0.5$.
(iii) Random Rotate: Randomly rotate the image from -45 degree to 45 degree.
(iv) Random Vertical Flip: Apply vertical flip over image with probability $p=0.5$.
(v) Random Horizontal Flip (BoostAdapter): Apply horizontal flip over image with probability $p=0.5$.
The results are presented in Table \ref{tab:aug_dg} and Table \ref{tab:aug_cross}. The results indicate that random horizontal flipping outperforms other augmentation methods, primarily because the images generated from horizontal flips are closer to the original distribution when training CLIP.

\begin{table}[!h]
    \caption{\textnormal{\textbf{Comparison of different augmentations on the OOD benchmark .} } Default settings are marked in \colorbox{baselinecolor}{gray}.}
    \small \centering
    \setlength{\tabcolsep}{8pt}
    \resizebox{\textwidth}{!}{
        \begin{tabular}{l|cccc|c}
            \toprule
                                               & Imagenet-V2    & Imagenet-Sketch & Imagenet-A      & Imagenet-R     & Average        \\
            \midrule
            Random Brightness                   &65.10	&51.24	&62.10	&\textbf{81.03}	&64.87         \\
            Random Auto Contrast              &65.50	&50.79	&64.33	&80.57	&65.30          \\
            Random Rotate                     &61.14	&47.67	&60.83	&78.15	&61.95         \\
            Random Vertical Flip              &63.39	&49.67	&60.77	&78.55	&63.10          \\
            \midrule
            \rowcolor{tabhighlight} Random Horizontal Flip &\textbf{65.51}	&\textbf{51.28}	&\textbf{64.53}	&{80.95}	&\textbf{65.57} \\
            \bottomrule
        \end{tabular}}
    \label{tab:aug_dg}
\end{table}

\begin{table}[!h]
    \caption{\textnormal{\textbf{Comparison of different augmentations on the Cross-Domain Benchmark.}} Default settings are marked in \colorbox{baselinecolor}{gray}.}
    \tabstyle{4pt}
    \resizebox{\textwidth}{!}{
        \begin{tabular}{l|cccccccccc|c}
            \toprule
                                                           & \rotatebox{90}{Caltech} & \rotatebox{90}{Pets} & \rotatebox{90}{Cars} & \rotatebox{90}{Flowers} & \rotatebox{90}{Food101} & \rotatebox{90}{Aircraft} & \rotatebox{90}{SUN397} & \rotatebox{90}{DTD} & \rotatebox{90}{EuroSAT} & \rotatebox{90}{UCF101} & \rotatebox{90}{\emph{ Average}} \\
            \midrule
            Random Brightness                               &94.60	&\textbf{89.70}	&69.28	&71.70	&86.88	&26.67	&\textbf{68.24}	&45.57	&61.63	&71.45	&68.57                          \\
            Random Auto Contrast                           &94.48	&89.67	&69.33	&71.90	&\textbf{87.24}	&27.39	&68.16	&45.51	&61.67	&71.77	&\textbf{68.71}                  \\
            Random Rotate                                  &94.52	&89.59	&67.74	&71.30	&85.91	&24.27	&67.56	&45.45	&60.72	&70.66	&67.77                           \\
            Random Vertical Flip                           & \textbf{94.89}                   & 89.53                & 68.75                & \textbf{72.19}                   & 86.78                   & 24.99                    & 67.72                  & 45.27               & \textbf{61.56}                   & 70.82                  & 68.25                           \\
            \rowcolor{tabhighlight} Random Horizontal Flip  &{94.77}	&89.51	&\textbf{69.30}	&{71.66}	&{87.17}	&\textbf{27.45}	&{68.09}	&\textbf{45.69}	&{61.22}	&\textbf{71.93}	&\textbf{68.68}                  \\
            \bottomrule
        \end{tabular}
    }
    \label{tab:aug_cross}
\end{table}

\section{Additional Ablation Results}
\para{Historical Samples and Boosting Samples.}
\label{sec:boosting_more}
We provide more ablation results of the historical and boosting samples on the Cross-Dataset benchmark in Table \ref{tab:boosting_cross}. The observation is consistent with the results in Table \ref{tab:boosting}, showing that CLIP gains improvements from both historical and boosting samples. Furthermore, when applied to various downstream tasks, the importance of regional bootstrapping becomes more significant, as indicated by the gap between BoostAdapter and the variant that uses  boosting samples only.

\begin{table}[h]
    \caption{\textnormal{\textbf{Ablation study on historical samples
and boosting sample on the Cross-Domain Benchmark.}}}
    \tabstyle{4pt}
    \resizebox{\textwidth}{!}{
        \begin{tabular}{l|cccccccccc|c}
            \toprule
                                                           & \rotatebox{90}{Caltech} & \rotatebox{90}{Pets} & \rotatebox{90}{Cars} & \rotatebox{90}{Flowers} & \rotatebox{90}{Food101} & \rotatebox{90}{Aircraft} & \rotatebox{90}{SUN397} & \rotatebox{90}{DTD} & \rotatebox{90}{EuroSAT} & \rotatebox{90}{UCF101} & \rotatebox{90}{\emph{ Average}} \\
            \midrule
            CLIP & 93.35 & 88.25 & 65.48 & 67.44 & 83.65 & 23.67 & 62.59 & 44.27 & 42.01 & 65.13 & 63.58 \\
            Historical Samples &94.16	&{89.42}	&66.87	&\textbf{72.11}	&85.93	&24.69	&67.24	&44.80	&\textbf{61.85}	&69.81	&67.69\\
            Boosting Samples &94.32	&88.64	&68.38	&71.54	&87.12	&{27.30}	&67.42	&44.68	&45.93	&69.34	&66.47\\
            \rowcolor{tabhighlight} BoostAdapter &\textbf{94.77}	&\textbf{89.51}	&\textbf{69.30}	&{71.66}	&\textbf{87.17}	&\textbf{27.45}	&\textbf{68.09}	&\textbf{45.69}	&{61.22}	&\textbf{71.93}	&\textbf{68.68}             \\
            \bottomrule
        \end{tabular}
    }
    \label{tab:boosting_cross}
\end{table}

\para{Number of Augmented Views for Boosting Samples.} 
The complete results on the number of augmented views are presented in Table \ref{tab:views_dg} and Table \ref{tab:views_cross}.  With more augmented views, BoostAdapter is able to better extract the fine-grained information from the original test sample, achieving improved performance.

\begin{table}[!h]
    \caption{\textnormal{\textbf{Results of different views on the OOD benchmark.} }Default settings are marked in \colorbox{baselinecolor}{gray}.}
    \small \centering
    \setlength{\tabcolsep}{8pt}
    \resizebox{\textwidth}{!}{
        \begin{tabular}{l|cccc|c}
            \toprule
                                               & Imagenet-V2    & Imagenet-Sketch & Imagenet-A      & Imagenet-R     & Average        \\
            \midrule
            16 Views &79.41 &49.01 &62.08		&63.68		&63.54\\
            32 Views &80.32 &50.73 &63.22		&64.91		&64.80\\
            \rowcolor{tabhighlight} 64 Views &80.95 &51.28 &64.53		&65.51		&65.57 \\
            128 Views &80.95&51.91  &64.06		&65.27		&65.55\\
            \bottomrule
        \end{tabular}}
    \label{tab:views_dg}
\end{table}

\begin{table}[!h]
    \caption{\textnormal{\textbf{Results of different views on the Cross-Domain Benchmark.}} Default settings are marked in \colorbox{baselinecolor}{gray}.}
    \tabstyle{4pt}
    \resizebox{\textwidth}{!}{
        \begin{tabular}{l|cccccccccc|c}
            \toprule
                                                           & \rotatebox{90}{Caltech} & \rotatebox{90}{Pets} & \rotatebox{90}{Cars} & \rotatebox{90}{Flowers} & \rotatebox{90}{Food101} & \rotatebox{90}{Aircraft} & \rotatebox{90}{SUN397} & \rotatebox{90}{DTD} & \rotatebox{90}{EuroSAT} & \rotatebox{90}{UCF101} & \rotatebox{90}{\emph{ Average}} \\
            \midrule
            16 Views &93.95	&89.62	&68.06	&71.62	&86.76	&25.71	&67.33	&45.39	&62.07	&70.97	&68.15\\
            32 Views &94.48	&89.59	&69.07	&71.54	&87.01	&27.18	&67.97	&45.45	&61.22	&71.56	&68.51\\
            \rowcolor{tabhighlight} 64 Views &94.77	&89.51	&69.3	&71.66	&87.17	&27.45	&68.09	&45.69	&61.22	&71.93	&68.68                  \\
            128 Views &94.77	&89.62	&69.15	&71.34	&87.28	&27.15	&68.15	&45.86	&61.19	&71.87	&68.64\\
            \bottomrule
        \end{tabular}
    }
    \label{tab:views_cross}
\end{table}

\begin{table}[h]
    \caption{\textnormal{\textbf{Results of fixed shot capacity on the OOD benchmark.} }} 
    \small \centering
 \setlength{\tabcolsep}{8pt}
    \resizebox{\textwidth}{!}{
    \begin{tabular}{l|cccc|c}
    \toprule
    & Imagenet-V2 & Imagenet-Sketch & Imagenet-A &  Imagenet-R  & Average\\
    \midrule
    CLIP  & 60.86 & 46.09 & 47.87 & 73.98 & 57.20 \\
    CLIP+TPT \ & 64.35 & 47.94 & 54.77 & 77.06 & 60.81 \\
PromptAlign  & 65.29 & 50.23 & 59.37  & 79.33 & 63.55 \\
    TDA & 64.67 & {50.54} & 60.11  & 80.24 & 63.89  \\
    \midrule
    BoostAdapter-Fixed &\textbf{65.13}	&\textbf{50.66}	&63.96	&80.44	&65.05\\
    \rowcolor{tabhighlight} BoostAdapter-Search &\textbf{65.03 }	&\textbf{50.66}	&\textbf{64.27 }	&\textbf{80.64 }	&\textbf{65.15 } \\
    \bottomrule
    \end{tabular}}
    \label{tab:fix_cache_dg}
\end{table}

\begin{table}[h]
    \caption{\textnormal{\textbf{Results of fixed shot capacity on the Cross-Domain Benchmark.}}}
    \tabstyle{4pt}
    \resizebox{\textwidth}{!}{
        \begin{tabular}{l|cccccccccc|c}
            \toprule
                                                           & \rotatebox{90}{Caltech} & \rotatebox{90}{Pets} & \rotatebox{90}{Cars} & \rotatebox{90}{Flowers} & \rotatebox{90}{Food101} & \rotatebox{90}{Aircraft} & \rotatebox{90}{SUN397} & \rotatebox{90}{DTD} & \rotatebox{90}{EuroSAT} & \rotatebox{90}{UCF101} & \rotatebox{90}{\emph{ Average}} \\
            \midrule
                CLIP  & 93.35 & 88.25 & 65.48 & 67.44 & 83.65 & 23.67 & 62.59 & 44.27 & 42.01 & 65.13 & 63.58 \\
    CLIP+TPT   & 94.16 & 87.79 & 66.87 & 68.98 & 84.67 & 24.78 & 65.50 & 47.75 & 42.44 & 68.04 & 65.10 \\
    PromptAlign & 94.01 & \textbf{90.76} & 68.50 & \textbf{72.39} & {86.65} & 24.80 & {67.54} & 47.24 & 47.86 & 69.47 & 66.92 \\
    TDA & 94.24 & 88.63 & 67.28 & 71.42 & 86.14 & 23.91 & 67.62 & \textbf{47.40} & 58.00  & 70.66 & 67.53 \\
    \midrule
    BoostAdapter-Fixed &\textbf{94.77}	&88.85	&\textbf{69.30}	&\textbf{71.66}	&87.17	&27.00	&67.64	&44.33	&\textbf{61.22}	&69.73	&68.17 \\
            \rowcolor{tabhighlight} BoostAdapter-Search &\textbf{94.77}	&{89.51}	&\textbf{69.30}	&\textbf{71.66}	&\textbf{87.17}	&\textbf{27.45}	&\textbf{68.09}	&\textbf{45.69}	&{61.22}	&\textbf{71.93}	&\textbf{68.68}                    \\
            \bottomrule
        \end{tabular}
    }
    \label{tab:fix_cache_cross}
\end{table}
\para{Fixed shot capacity.} We search the optimal total shot capicity in BoostAdapter. We also find that fixing the cache size to be 3 can generalize well in different task settings,
as shown in Table \ref{tab:fix_cache_dg} and Table \ref{tab:fix_cache_cross}.

\section{More Qualitative Results}
More qualitative results are provided in Fig. \ref{fig:more_vis}. 
\begin{table}[!h]
\centering
    \includegraphics[width=\linewidth]{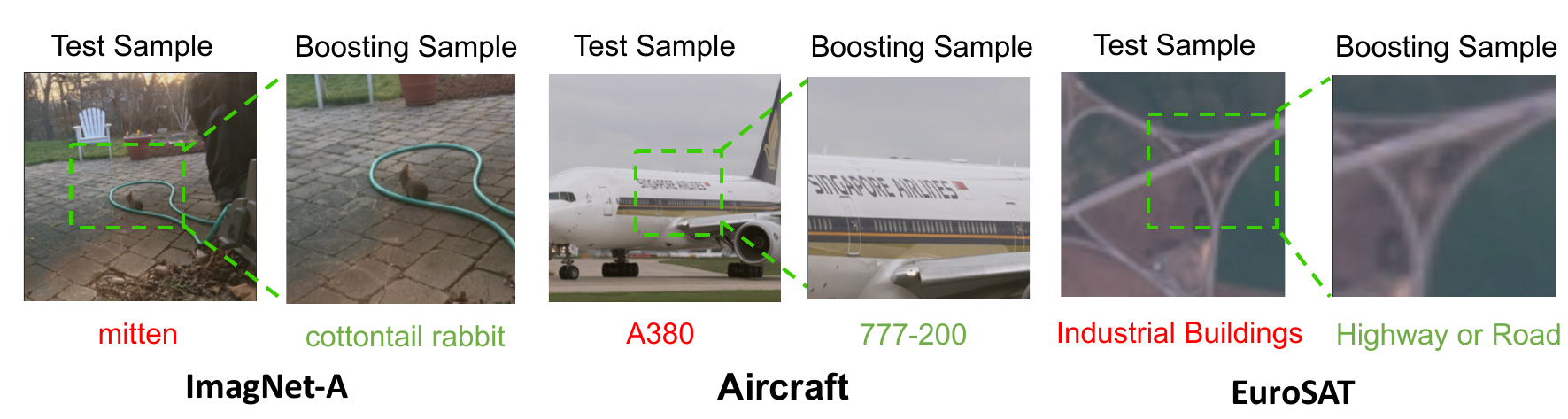}
    \captionof{figure}{More qualitative results on ImagNet-A, Aircraft and EuroSAT. }
    \label{fig:more_vis}
\end{table}
\clearpage
\newpage
\section*{NeurIPS Paper Checklist}

\begin{enumerate}

\item {\bf Claims}
    \item[] Question: Do the main claims made in the abstract and introduction accurately reflect the paper's contributions and scope?
    \item[] Answer: \answerYes{} 
    \item[] Justification: 
    The main claims made in the abstract and introduction reflect the main idea described in  Section \ref{sec:method}. 

\item {\bf Limitations}
    \item[] Question: Does the paper discuss the limitations of the work performed by the authors?
    \item[] Answer: \answerYes{} 
    \item[] Justification: 
    The limitation and the discussion about computational overhead can be found in the Section \ref{sec:conslusions}. These assumptions are reasonable in domain adaptation and parameter analysis is conduct in the ablation studies in Section \ref{sec:ablation}.

\item {\bf Theory Assumptions and Proofs}
    \item[] Question: For each theoretical result, does the paper provide the full set of assumptions and a complete (and correct) proof?
    \item[] Answer: \answerYes{} 
    \item[] Justification: 
    The theoretical proof of the proposition used in the paper can be found in Section \ref{sec:proof}.

    \item {\bf Experimental Result Reproducibility}
    \item[] Question: Does the paper fully disclose all the information needed to reproduce the main experimental results of the paper to the extent that it affects the main claims and/or conclusions of the paper (regardless of whether the code and data are provided or not)?
    \item[] Answer: \answerYes{} 
    \item[] Justification: 
    We provide the implementation details in Section \ref{sec:details} for reproduction.

\item {\bf Open access to data and code}
    \item[] Question: Does the paper provide open access to the data and code, with sufficient instructions to faithfully reproduce the main experimental results, as described in supplemental material?
    \item[] Answer: \answerYes{} 
    \item[] Justification: 
    We will open source the code once accepted.

\item {\bf Experimental Setting/Details}
    \item[] Question: Does the paper specify all the training and test details (e.g., data splits, hyperparameters, how they were chosen, type of optimizer, etc.) necessary to understand the results?
    \item[] Answer: \answerYes{} 
    \item[] Justification: 
    We provide the implementation details in Section \ref{sec:details}.

\item {\bf Experiment Statistical Significance}
    \item[] Question: Does the paper report error bars suitably and correctly defined or other appropriate information about the statistical significance of the experiments?
    \item[] Answer: \answerYes{} 
    \item[] Justification:
    We provide the error bound along with the main results.

\item {\bf Experiments Compute Resources}
    \item[] Question: For each experiment, does the paper provide sufficient information on the computer resources (type of compute workers, memory, time of execution) needed to reproduce the experiments?
    \item[] Answer: \answerYes{} 
    \item[] Justification: 
    We provide the information about compute resources in the implementation details in Section \ref{sec:details}.
    
\item {\bf Code Of Ethics}
    \item[] Question: Does the research conducted in the paper conform, in every respect, with the NeurIPS Code of Ethics \url{https://neurips.cc/public/EthicsGuidelines}?
    \item[] Answer: \answerYes{} 
    \item[] Justification: 
    The research conducted in the paper conform with the NeurIPS Code of Ethics. 

\item {\bf Broader Impacts}
    \item[] Question: Does the paper discuss both potential positive societal impacts and negative societal impacts of the work performed?
    \item[] Answer: \answerYes{} 
    \item[] Justification: 
    We provide discussions of broader impacts in Section \ref{sec:impacts} in Appendix.
    
\item {\bf Safeguards}
    \item[] Question: Does the paper describe safeguards that have been put in place for responsible release of data or models that have a high risk for misuse (e.g., pretrained language models, image generators, or scraped datasets)?
    \item[] Answer: \answerNA{} 
    \item[] Justification: \answerNA{}

\item {\bf Licenses for existing assets}
    \item[] Question: Are the creators or original owners of assets (e.g., code, data, models), used in the paper, properly credited and are the license and terms of use explicitly mentioned and properly respected?
    \item[] Answer: \answerYes{} 
    \item[] Justification: 
    We mention the licenses of existing assets in the Section \ref{sec:licenses} in Appendix.

\item {\bf New Assets}
    \item[] Question: Are new assets introduced in the paper well documented and is the documentation provided alongside the assets?
    \item[] Answer: \answerNA{} 
    \item[] Justification: \answerNA{}

\item {\bf Crowdsourcing and Research with Human Subjects}
    \item[] Question: For crowdsourcing experiments and research with human subjects, does the paper include the full text of instructions given to participants and screenshots, if applicable, as well as details about compensation (if any)? 
    \item[] Answer: \answerNA{} 
    \item[] Justification: \answerNA{}

\item {\bf Institutional Review Board (IRB) Approvals or Equivalent for Research with Human Subjects}
    \item[] Question: Does the paper describe potential risks incurred by study participants, whether such risks were disclosed to the subjects, and whether Institutional Review Board (IRB) approvals (or an equivalent approval/review based on the requirements of your country or institution) were obtained?
    \item[] Answer: \answerNA{} 
    \item[] Justification: \answerNA{}

\end{enumerate}

\end{document}